%% file: main.tex
\documentclass[10pt,twocolumn,letterpaper]{article}

\usepackage{cvpr} 

\definecolor{cvprblue}{rgb}{0.21,0.49,0.74}
\usepackage[pagebackref,breaklinks,colorlinks,allcolors=cvprblue]{hyperref}
\usepackage{array}
\usepackage{multirow}

\title{Enhancing Diffusion-based Restoration Models via Difficulty-Adaptive \\ Reinforcement Learning with IQA Reward}

\author{ Xiaogang Xu$^{1}$ \quad Ruihang Chu$^{2}$ \quad Jian Wang$^{3}$ \quad Kun Zhou$^{4}$ \\ \quad Wenjie Shu$^{5}$ \quad Harry Yang$^{5}$ \quad Ser-Nam Lim$^{6}$  \quad Hao Chen$^{7}$ \quad Liang Lin$^{8}$ \\
	$^1$ The Chinese University of Hong Kong
	\quad $^2$ Tsinghua University 
	\quad $^3$ Snap Research \\
	\quad $^4$ Shenzhen University
	\quad $^5$ HKUST  \quad $^6$ University of Central Florida \\
	 \quad $^7$ UC Davis \quad  $^8$ Sun Yat-Sen University \\
	{\tt \small xiaogangxu00@gmail.com} \\
}

\begin{document}
\maketitle
\input{sec/0_abstract}    
\input{sec/1_intro}

\input{sec/2_related_work}

\input{sec/3_method}

\input{sec/4_experiment}

\input{sec/5_conclusion}

{
    \small
    \bibliographystyle{ieeenat_fullname}
    \bibliography{main}
}

\end{document}

%% file: sec/0_abstract.tex
\begin{abstract}
Reinforcement Learning (RL) has recently been incorporated into diffusion models, e.g., tasks such as text-to-image. However, directly applying existing RL methods to diffusion-based image restoration models is suboptimal, as the objective of restoration fundamentally differs from that of pure generation: it places greater emphasis on fidelity.
In this paper, we investigate how to effectively integrate RL into diffusion-based restoration models.
First, through extensive experiments with various reward functions, we find that an effective reward can be derived from an Image Quality Assessment (IQA) model, instead of intuitive ground-truth-based supervision, which has already been optimized during the Supervised Fine-Tuning (SFT) stage prior to RL.
Moreover, our strategy focuses on using RL for challenging samples that are significantly distant from the ground truth, and our RL approach is innovatively implemented using MLLM-based IQA models to align distributions with high-quality images initially. As the samples approach the ground truth's distribution, RL is adaptively combined with SFT for more fine-grained alignment. This dynamic process is facilitated through an automatic weighting strategy that adjusts based on the relative difficulty of the training samples.
Our strategy is plug-and-play that can be seamlessly applied to diffusion-based restoration models, boosting its performance across various restoration tasks.
Extensive experiments across multiple benchmarks demonstrate the effectiveness of our proposed RL framework.
\end{abstract}

%% file: sec/1_intro.tex
\section{Introduction}
\label{sec:intro}

Diffusion-based generative restoration~\cite{wang2024exploiting,yang2024pixel,lin2024diffbir} has attracted increasing attention for its strong ability to synthesize photo-realistic content from severe degradations.
This approach typically involves augmenting large pre-trained text-to-image models~\cite{rombach2022high,flux2024} with a control branch~\cite{zhang2023adding}, employing Supervised FineTuning (SFT).
Compared with traditional image restoration methods~\cite{xu2022snr,zamir2022restormer}, diffusion-based approaches tend to produce outputs that are sharper and cleaner. 
However, hallucination also occurs since the generated content may include elements that do not exist in the original Low-Quality (LQ) image~\cite{pan2025boosting,zhang2024diffusion}, and some unnatural textures and colors may exist.
This may be because SFT focuses solely on reference-based alignment, while most restoration tasks are ill-posed, making direct optimization toward references (ground truths) suboptimal.

\begin{figure}[t]
	\begin{center}
		\includegraphics[width=1.0\linewidth]{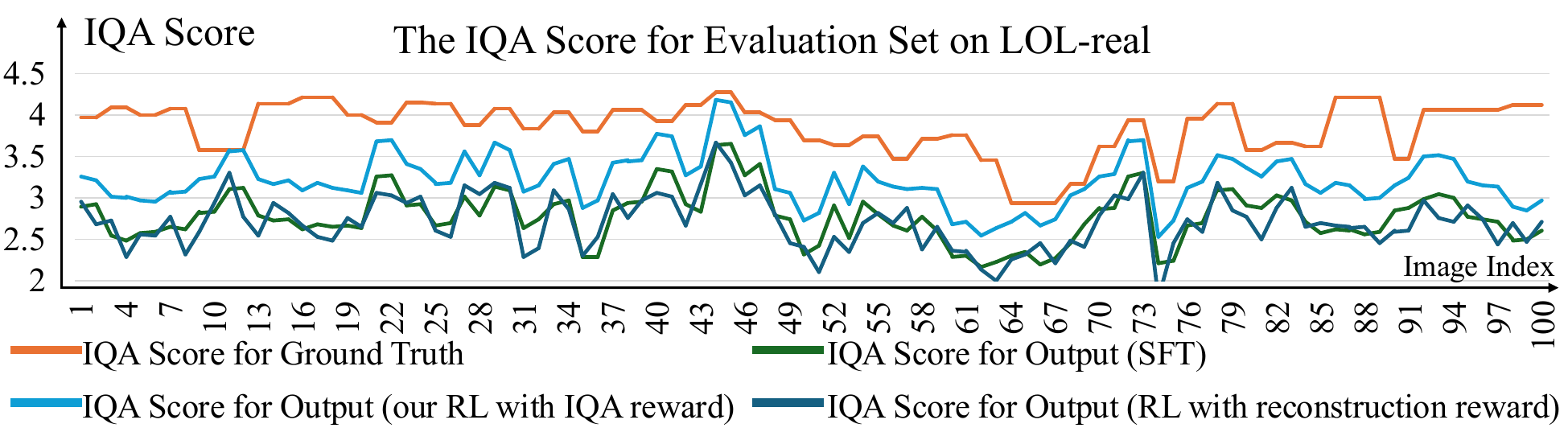}
	\end{center}
    \vspace{-0.2in}
    \caption{
IQA scores in this figure are computed using the MLLM-based IQA model DeQA-Score~\cite{you2025teaching} on the widely-used real-world low-light image enhancement benchmark LOL-real~\cite{yang2021sparse}.
The original IQA score of the diffusion-based method (DiffBIR~\cite{lin2024diffbir}) via SFT shows a significant gap compared to the IQA score of the ground truth. 
Thus, IQA can serve to distinguish the distribution between output images and ground truths.
After our RL training, it aligns more closely with the ground-truth distribution.
	}
	\label{fig:teaser}
    \vspace{-0.2in}
\end{figure}

\begin{figure}[t]
    \begin{center}
		\includegraphics[width=1.0\linewidth]{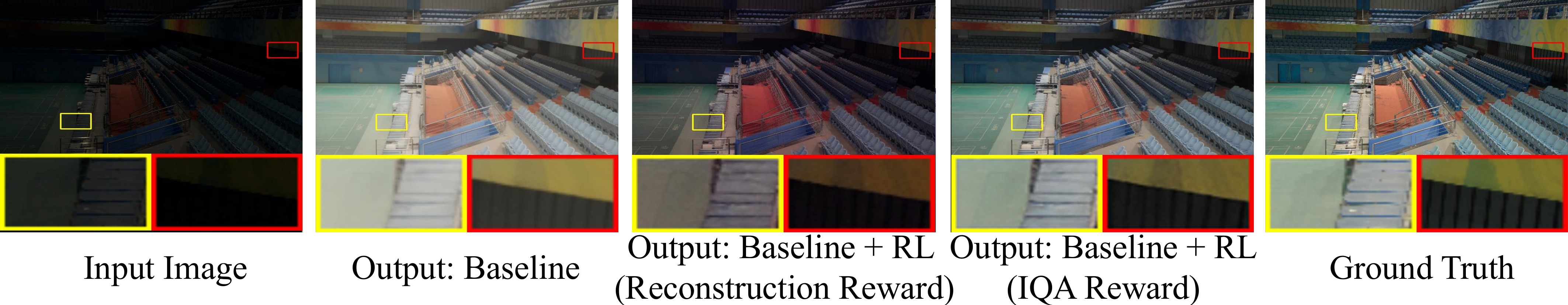}
	\end{center}
    \vspace{-0.2in}
    \caption{Visual comparisons using reconstruction error and IQA as the reward function. The reward guided by IQA leads to better visual performance, producing results that are closer to the ground truth, whereas the reconstruction-error-based reward fails.
	}
	\label{fig:reward-visual}
    \vspace{-0.22in}
\end{figure}

\begin{figure*}[t]
	\begin{center}
		\includegraphics[width=1.0\linewidth]{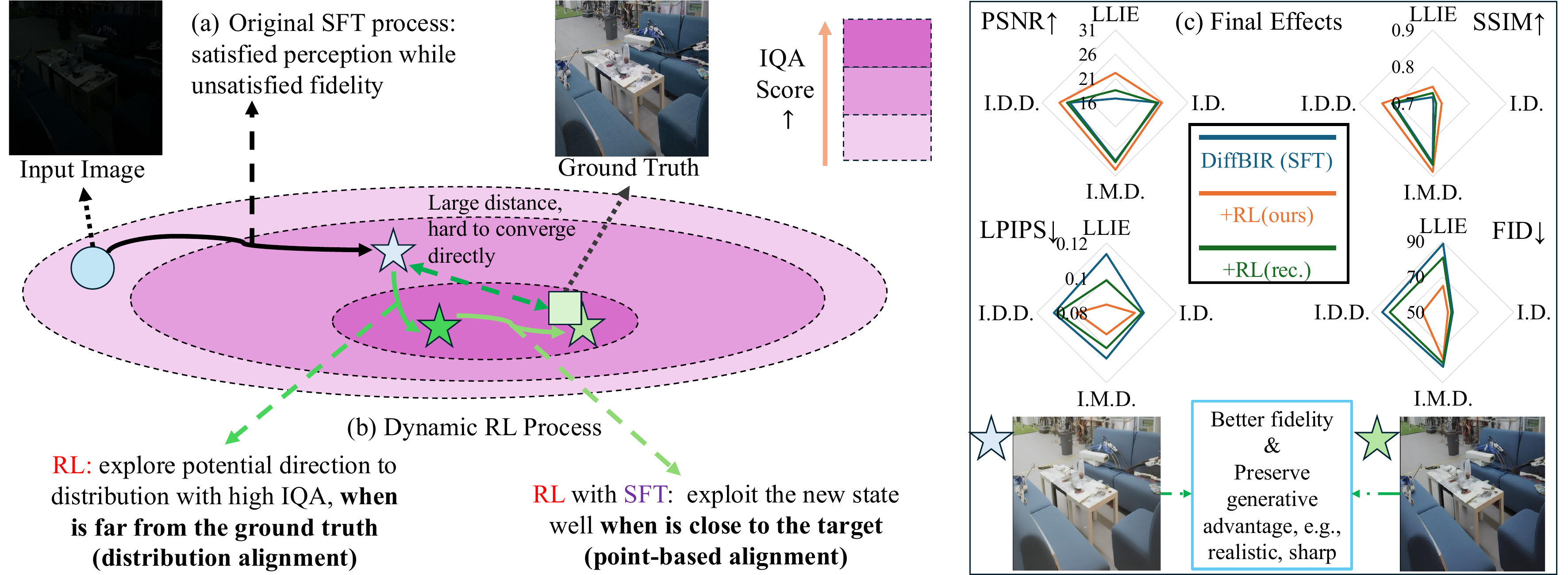}
	\end{center}
    \vspace{-0.2in}
    \caption{
The schematic diagram of our RL algorithm (the stars $\bigstar$ of various colors in (a) and (b) represent the change of sample's latent state during training).
(a) Although diffusion-based restoration models via SFT can generate samples with satisfactory perception, there still exist challenging cases far away from ground truths. 
(b) Our proposed RL process is dynamic: guided by IQA-based rewards, focuses on these underperforming samples, first exploring potential optimization directions and gradually aligning with the target distribution. Once the results approach the target distribution, SFT is combined to ensure fine-grained reference-based alignment.
(c) Results optimized via RL not only demonstrate improved fidelity but also retain generative advantages of diffusion-based models. The radar map shows the quantitative comparisons between the current SOTA diffusion-based method (e.g., DiffBIR) via SFT and its variants with our RL strategy across multiple datasets, including LOL-real (low-light image enhancement, LLIE), Rain100H (image deraining, I.D.), GoPro (image motion deblurring, I.M.D.), and DPDD (image defocus deblurring, I.D.D.). More results are provided in the experimental section.
	}
	\label{fig:basic}
    \vspace{-0.2in}
\end{figure*}

On the other hand, Reinforcement Learning (RL) has recently proven effective in behavior alignment of large models (such as text-to-image models~\cite{black2023training, fan2023dpok, miao2024training, zhang2024large}) after SFT. 
To the best of our knowledge, it has not been applied to restoration tasks with large diffusion models. Here, we make the first attempt.
The restoration task has a well-defined target: the high-quality ground truth~\cite{xu2022snr,xu2023low,xu2022deep,wang2021seeing}. 
A straightforward approach is to use the distance between the model output and the ground truth as the reward.
However, we observe that this type of reward tends to have limited effectiveness.
There may be two reasons. First, the SFT stage before RL already optimizes a distance-based objective. As a result, some samples may follow the same search direction established during SFT and fail to explore potentially better solutions. 
This is reflected in the similar IQA scores for RL with reconstruction reward compared to the baselines (Fig.~\ref{fig:teaser}) and unstable reward curve (Fig.~\ref{fig:reward}).
Moreover, it's still a reference-based reward that is highly ill-posed to be optimize, and even produce worse results (Fig.~\ref{fig:reward-visual}).

In this paper, we propose a novel RL strategy tailored for diffusion-based restoration.
We demonstrate that RL can be effectively applied to diffusion-based restoration by using Multi-modal Large Language Models (MLLM)-based IQA models~\cite{you2025teaching} as reward functions and adopting a difficulty-adaptive training strategy. 
Such a RL strategy plays two roles: 1) first guide the diffusion model toward alternative solutions to achieve distribution level alignment (exploration); 2) then enable the model exploit these new directions to better align with the target, collaborating with SFT (exploitation).
This is similar to the iterative training of RL and SFT in DeepSeek~\cite{guo2025deepseek}, while it is grounded in a different theoretical framework that is more suitable for diffusion-based restoration networks.

\noindent\textbf{Distribution-level alignment.} Modern MLLM-based methods can effectively regress accurate quality scores by leveraging their broad training and inherent adaptability~\cite{you2025teaching,qalign}. 
Our key insight is that restored images with both high fidelity and realism should receive higher IQA scores compared to those containing fidelity-related artifacts~\cite{wang2024exploiting,wu2024one}. This is because IQA models are trained on real images and are inherently biased toward content that aligns with physical plausibility and human perception (one example for the IQA scores of real and generated outputs is shown in Fig.~\ref{fig:teaser}). 
Thus, \textit{guiding the generation process toward outputs with higher IQA scores naturally steers the model's distribution toward the ground truth}, i.e., the proposed reward encourages alignment at the distribution level.

\noindent\textbf{Fine-grained reference-based alignment.} RL with distribution alignment may fall short of achieving full alignment with ground truths.
To address this, we propose a difficulty-adaptive training strategy that combines the strengths of both RL and SFT. 
We apply IQA-based RL strategy primarily to hard samples (those with large discrepancies from the ground truth), where traditional SFT struggles to make progress. For such samples, RL can effectively help discover alternative generation paths that direct the model toward distributions of improved fidelity and realism.
Once the distribution of RL-trained samples becomes sufficiently close to the target, we progressively reintroduce SFT to further refine the output and bring it closer to the ground truth via reference-based alignment. 
This mechanism can be viewed in Fig.~\ref{fig:basic}.
This adaptive switching is achieved via an automatic mechanism that assigns dynamic loss weights to individual samples based on difficulty levels.

Furthermore, we identify several key techniques that enhance RL in restoration tasks. First, we observe that policy modeling can benefit from a reliable direction by using a better denoised latent as the target for the action, helping the model learn more meaningful updates.
Additionally, our RL strategy differs from prior methods for text-to-image~\cite{black2023training,liu2025flow}, which typically apply the reward function only to the output at the final diffusion step. In contrast, we apply RL supervision at each intermediate step in the diffusion process. This step-wise reward application helps mitigate error accumulation and provides more consistent guidance throughout the denoising trajectory. 

We conduct extensive experiments across multiple datasets to validate the effectiveness of our strategy.
Our method is general and plug-and-play, making it applicable to existing diffusion-based generative restoration networks.
In summary, our contributions are three-fold:
\begin{itemize}
    \item We are the first to propose an effective RL training pipeline tailored for restoration-targeted diffusion models, highlighting key implementation techniques.

    \item We highlight the impact of incorporating an IQA-based reward (not explored in previous studies) and propose a difficulty-adaptive strategy within the RL to better accommodate the unique property of restoration tasks.

    \item We conduct extensive experiments on public datasets to show the effectiveness of our RL strategy in improving the performance of current diffusion-based methods.

\end{itemize}

\noindent\textbf{Note} that our method is primarily designed for \textit{diffusion models built upon large pre-trained base models (e.g., text-to-image models such as Stable Diffusion and Flux), as these models often face significant challenges in preserving fidelity and avoiding artifacts}.

%% file: sec/2_related_work.tex
\section{Related Works}
\label{sec:related_works}

\noindent\textbf{Restoration Models using Pre-trained Diffusion.}
With the advancement of pre-trained diffusion models~\cite{rombach2022high,flux2024}, image restoration tasks have opened up new opportunities. 
Recent efforts in image super-resolution~\cite{wang2024exploiting,yang2024pixel,lin2024diffbir,wu2024one,sun2024coser,yu2024scaling,qu2024xpsr,sun2024pixel,chen2024faithdiff,luo2024photo,wang2024rap,dong2024tsd,qu2024xpsr} have largely focused on developing control mechanisms. 
StableSR~\cite{wang2024exploiting} and PASD~\cite{yang2024pixel} marks the initial attempts to implement the pre-trained diffusion strategy, utilizing the ControlNet.
DiffBIR~\cite{lin2024diffbir} further adds the region-adaptive restoration guidance, which can modify the denoising process during inference to enhance fidelity. 
Despite these advancements, aligning with ground truths remains a significant challenge.

\noindent\textbf{Reinforcement Learning for Diffusion Models.}
There have been several attempts to modify the biases of diffusion models using reinforcement learning~\cite{black2023training, fan2023dpok, miao2024training, zhang2024large,liu2025flow}. 
Denoising Diffusion Policy Optimization (DDPO)~\cite{black2023training} is an RL-based approach that reframes the diffusion process as a multi-step Markov Decision Process (MDP) to optimize a given reward function. 
Later, the effectiveness of various reward functions has been demonstrated in the text-to-image generation task, including diversity-based rewards\cite{zhang2024large, miao2024training}, alignment rewards~\cite{black2023training}, and visual rewards like aesthetic quality~\cite{black2023training}.
However, their applicability in image restoration remains underexplored.

\noindent\textbf{MLLM-based IQA Models.}
MLLM-based IQA methods leverage the foundational knowledge of MLLMs to achieve better performance~\cite{iqasurvey_tianhe, iqasurvey_zicheng}. 
E.g., Q-Bench~\cite{qbench, qbench_plus} proposes a binary softmax strategy that enables MLLMs to generate quality scores by predicting two discrete quality levels. 
Compare2Score~\cite{compare2score} achieves quality scoring by training MLLMs to compare pairs of images. Inspired by human annotation processes, Q-Align~\cite{qalign} discretizes scores into five discrete levels using one-hot labels to train MLLMs, resulting in more accurate score regression. Dog-IQA~\cite{dogiqa} uses a one-hot label for training-free IQA, incorporating specific standards and local semantic objects. DeQA-Score~\cite{you2025teaching} employs a distribution-based approach that discretizes the score distribution into soft labels, consistently outperforming other methods in score regression.

%% file: sec/3_method.tex
\begin{figure*}[t]
	\begin{center}
		\includegraphics[width=0.8\linewidth]{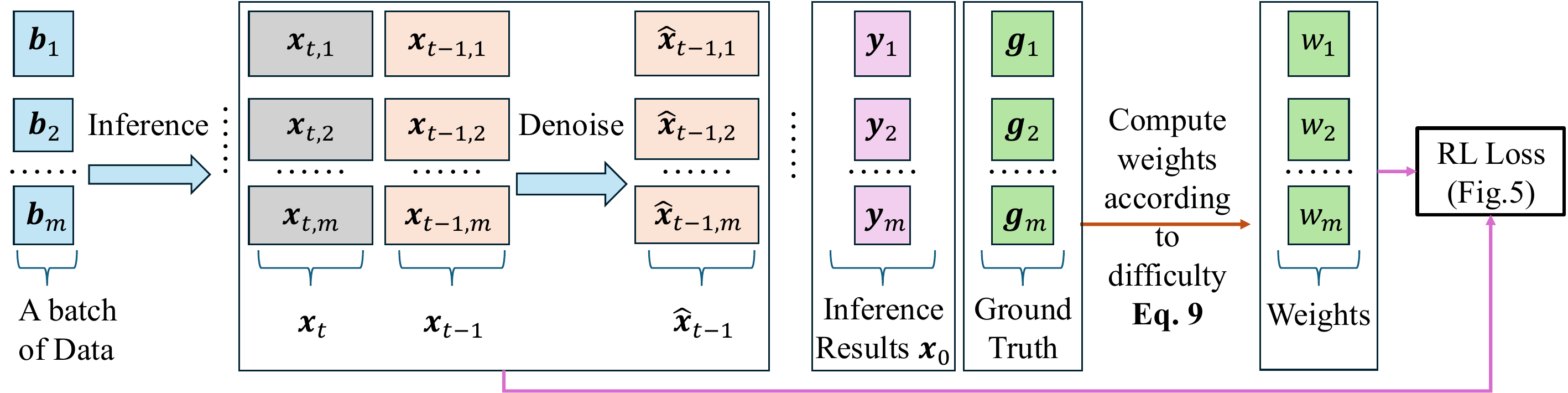}
	\end{center}
    \vspace{-0.25in}
    \caption{The illustration of our RL process, especially including the policy modeling and difficulty weights computation. The computation of RL loss can be viewed in Fig.~\ref{fig:pipeline2}.
	}
	\label{fig:pipeline}
    \vspace{-0.2in}
\end{figure*}

\section{Method}
\label{sec:method}

\subsection{Applying Reinforcement Learning to Diffusion}
\label{sec:diffusion}

\noindent\textbf{Diffusion models:}  Markov chains~\cite{sohl2015deep} model the data generation process by gradually adding and removing noise. The forward process transforms input data \( \boldsymbol{x}_0 \) into Gaussian noise \( \boldsymbol{x}_T \) over \( T \) steps using a variance schedule \( \beta_t \):
\begin{equation}
\begin{aligned}
    &q(\boldsymbol{x}_t | \boldsymbol{x}_{t-1}) = \boldsymbol{\mathcal{N}}(\sqrt{1 - \beta_t} \boldsymbol{x}_{t-1}, \beta_t \boldsymbol{I}),\\
    &q(\boldsymbol{x}_t | \boldsymbol{x}_0) = \boldsymbol{\mathcal{N}}(\sqrt{\bar{\alpha_t}} \boldsymbol{x}_0, (1 - \bar{\alpha_t}) \boldsymbol{I}),
\end{aligned}
\end{equation}
where \( \alpha_t = 1 - \beta_t \) and \( \bar{\alpha_t} = \prod_{s=1}^{t} \alpha_s \).
The reverse diffusion process in DDPMs \cite{ho2020denoising} removes noise using a trained model \( p_{\boldsymbol{\theta}}(\boldsymbol{x}_{t-1} | \boldsymbol{x}_t) \), approximating the true posterior \( q(\boldsymbol{x}_{t-1} | \boldsymbol{x}_t, \boldsymbol{x}_0) \). The model is trained via variational inference by minimizing KL divergence between $p$ and $q$.

\noindent\textbf{Diffusion with RL:} The diffusion model can be be further optimized to maximize the expected reward 
\begin{equation}
J(\boldsymbol{\theta})=\mathbb{E}_{\boldsymbol{c}\sim p(\boldsymbol{c}), \boldsymbol{x}\sim p_{\boldsymbol{\theta}}(\boldsymbol{x}|\boldsymbol{c})}[r(\boldsymbol{x}, \boldsymbol{c})],
\label{eq:policy-gradient}
\end{equation}
where $p(\boldsymbol{c})$ is distribution over input conditions (LQ image in the restoration scenario) and $p_{\boldsymbol{\theta}}(\boldsymbol{x}_0|\boldsymbol{c})$ is the sample distribution. This is achieved by first re-framing the diffusion model as a multi-step Markov Decision Process (MDP), containing a set of states S, actions A, reward function R, and state transition distribution P.
RL tries to maximize the reward function by learning the policy $\pi(\boldsymbol{a}_t|\boldsymbol{s}_t)$.
Current diffusion-based RL methods~\cite{black2023training,liu2025flow} often define MDP by using the denoising model backward process as the policy:
\begin{equation}
\begin{aligned}
    &\boldsymbol{s}_t \triangleq \{\boldsymbol{x}_t, \boldsymbol{c}, t\}, \; \boldsymbol{a}_t \triangleq \hat{\boldsymbol{x}}_{t-1}, \; \pi(\boldsymbol{a}_t | \boldsymbol{s}_t) \triangleq p_{\boldsymbol{\theta}}(\hat{\boldsymbol{x}}_{t-1} | \boldsymbol{x}_t, \boldsymbol{c}), \\
    &P(\boldsymbol{s}_{t+1} | \boldsymbol{s}_t, \boldsymbol{a}_t) \triangleq \{\delta(\hat{\boldsymbol{x}}_{t-1}), \delta(\boldsymbol{c}), \delta(t - 1)\}, \\
    &R(\boldsymbol{s}_t, \boldsymbol{a}_t) \triangleq r(\hat{\boldsymbol{x}}_{t-1}, \boldsymbol{c}),
\end{aligned}
\label{ex:state}
\end{equation}
where the state $\boldsymbol{s}_t$ is defined as the combination of latent $\boldsymbol{x}_t$, time $t$, and the condition $\boldsymbol{c}$. Policy $\pi$ for selecting the action is defined using the denoising model, $\delta(\boldsymbol{x})$ denotes the Dirac function, and $\hat{\boldsymbol{x}}_{t-1}$ is the denoised version of $\boldsymbol{x}_{t-1}$ via the diffusion network.
While fine-tuning, we use the importance sampling estimator~\cite{kakade2002approximately} which uses two models $\boldsymbol{\theta}$ and $\boldsymbol{\theta}_{\text{old}}$ to calculate the policy gradients for $\boldsymbol{\theta}$: 
\begin{equation}
\label{eq:clip_obj}
\resizebox{0.87\linewidth}{!}{$J(\boldsymbol{{\theta}}) = \mathbb{E} \left[ \sum_{t=0}^{T} \min \left[ w({ \boldsymbol{\theta}}, { \boldsymbol{\theta}_\text{old}})
 \hat{A}(\hat{\boldsymbol{x}}_{t-1}, \boldsymbol{c}), g(\epsilon,\hat{A}(\hat{\boldsymbol{x}}_{t-1},\boldsymbol{c})) \right] \right]$},
\end{equation}
where
\begin{gather*}
\resizebox{0.87\linewidth}{!}{$w(\boldsymbol{\theta}, \boldsymbol{ \theta}_\text{old}) = \frac{p_{\boldsymbol{\theta}}(\hat{\boldsymbol{x}}_{t-1}|\boldsymbol{x}_t,\boldsymbol{c})} {p_{\boldsymbol{\theta}_{\text{old}}}(\hat{\boldsymbol{x}}_{t-1}|\boldsymbol{x}_t,\boldsymbol{c})},
\quad g(\epsilon,\hat{A}) =
\begin{cases}
      (1+\epsilon)\hat{A} & \text{if $\hat{A}\geq0$}\\
      (1-\epsilon)\hat{A} & \text{if $\hat{A}<0$}
    \end{cases}$},
\end{gather*}
where $\epsilon$ is the hyper-parameter for clip interval, and $\hat{A}(\hat{\boldsymbol{x}}_{t-1}, {c})$ is the estimated advantage. 
$\hat{A}$ is actually the normalized award, typically with zero mean and unit variance to increase training stability. 
In policy-based RL, a general approach is to subtract a baseline state value function from the reward to obtain the \textit{advantage function}~\cite{NIPS1999_464d828b}
\begin{equation}
\hat{A}(\hat{\boldsymbol{x}}_{t-1}, \boldsymbol{c}) = (r(\hat{\boldsymbol{x}}_{t-1}, \boldsymbol{c})-\mu_r)/\sqrt{\sigma_{r}^2+\epsilon}.
\end{equation} 
Current diffusion-based RL normalizes rewards on a per-context basis by keeping track of a running mean and standard deviation for each prompt~\cite{black2023training}. 
Instead, we compute the mean and variance using both the track of rewards for each input image and the reward values from the current batch.

\begin{figure*}[t]
    \begin{center}
		\includegraphics[width=0.8\linewidth]{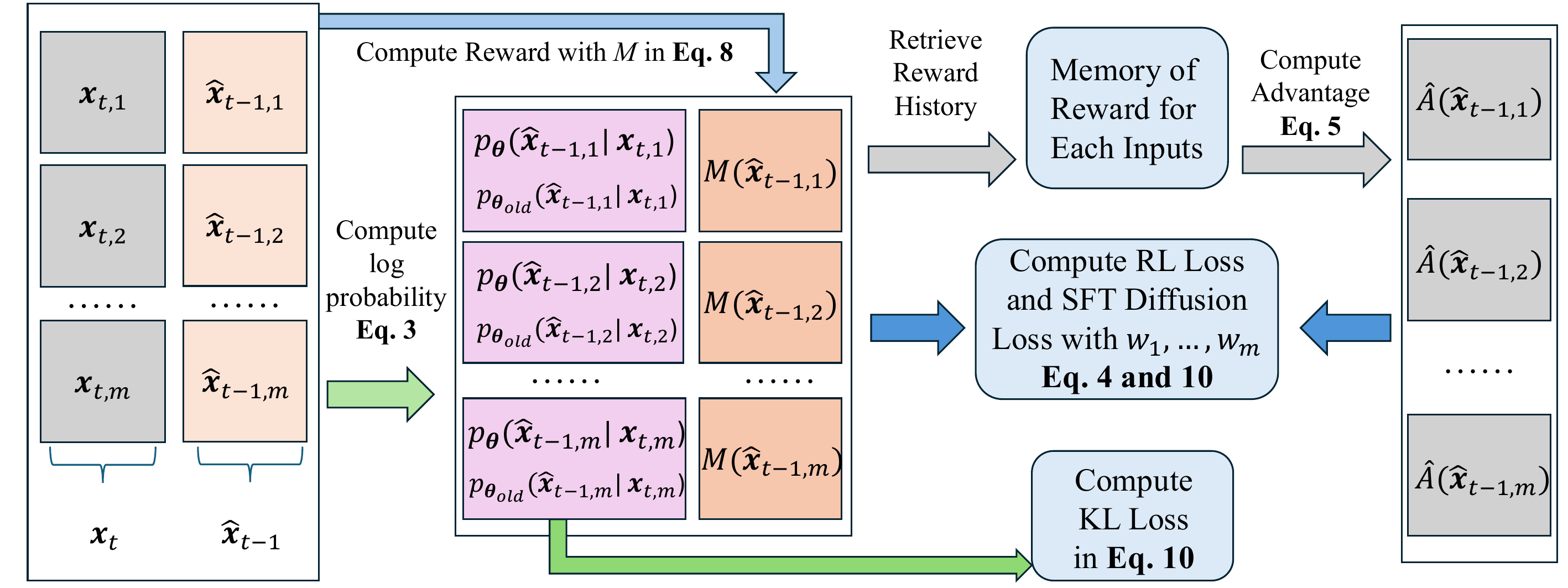}
	\end{center}
    \vspace{-0.25in}
    \caption{The illustration of our RL process that compute the loss on multiple time steps with reward normalization and KL constraint.
	}
	\label{fig:pipeline2}
    \vspace{-0.2in}
\end{figure*}

\textbf{Note that}, there are some modeling differences with current diffusion-based RL methods~\cite{black2023training,liu2025flow}, in terms of Eq.~\ref{ex:state}.

\noindent\textbf{Policy Modeling with A Better Direction.}
In the current diffusion-based RL, the policy model selects an action by denoising $\boldsymbol{x}_t$ to obtain the next state $\boldsymbol{x}_{t-1}$.
We observe that this process relates closely to the concept of exploration in RL—better actions correspond to better denoised results. In other words, we can improve the policy by providing a refined denoised result $\hat{\boldsymbol{x}}_{t-1}$ that is closer to the clean image $\boldsymbol{x}_0$. In this paper, we achieve this by applying an additional denoising step to $\boldsymbol{x}_t$, producing an enhanced estimate $\hat{\boldsymbol{x}}_{t-1}$. 

\noindent\textbf{Reward at Different Time Steps of Diffusion Generation.}
We observe that the current diffusion-based RL methods apply the reward only to the final output $\boldsymbol{x}_0$.
However, we argue that this strategy is not well-suited for restoration tasks. In diffusion-based restoration models, intermediate results contain important signals related to fidelity. In other words, if fidelity degradation occurs at intermediate time steps during the diffusion process, the final output $\boldsymbol{x}_0$ is likely to suffer from fidelity issues as well. Therefore, we propose to apply RL at each time step in the diffusion model, aiming to prevent artifact accumulation and fidelity loss throughout the denoising trajectory.

\subsection{Reward for Diffusion-based RL in Restoration}
\label{sec:reward}

A well-designed reward should guide the network in balancing exploration and exploitation within RL. We assume that exploration involves seeking additional possible solutions and information by trying different actions beyond the current choice, while exploitation focuses on leveraging known actions and information to obtain better results.

\noindent\textbf{Reward with GT supervision.}
Unlike generative tasks such as text-to-image, generative restoration tasks impose strict requirements on closing to the ground truths.
Specifically, the quality of the generated images must be adapted to the target domain (e.g., from low-light to normal-light~\cite{xu2022snr,cai2023retinexformer}), while preserving fidelity.
Therefore, a straightforward reward function approach is to measure the similarity between the output and the high-quality ground truth, as
\begin{equation}
    r(\hat{\boldsymbol{x}}_{t-1}, \boldsymbol{c})=\Vert \hat{\boldsymbol{x}}_{t-1} - \boldsymbol{g} \Vert,
\end{equation}
where $\boldsymbol{g}$ represents the ground truth.
However, after extensive experimentation, we found that such a reference-based reward model does not yield positive effects. We think this is first because RL is applied to a SFT-trained diffusion model, whose training objective already minimizes the distance between the output and the ground truth~\cite{wang2024exploiting}. In other words, the guidance from RL aligns with the objective of SFT, offering little additional benefits. 
Moreover, closing to a point in the latent space is highly ill-posed, which is hard and unstable to optimize and even has worse results.

\begin{figure}[t]
    \begin{center}
		\includegraphics[width=1.0\linewidth]{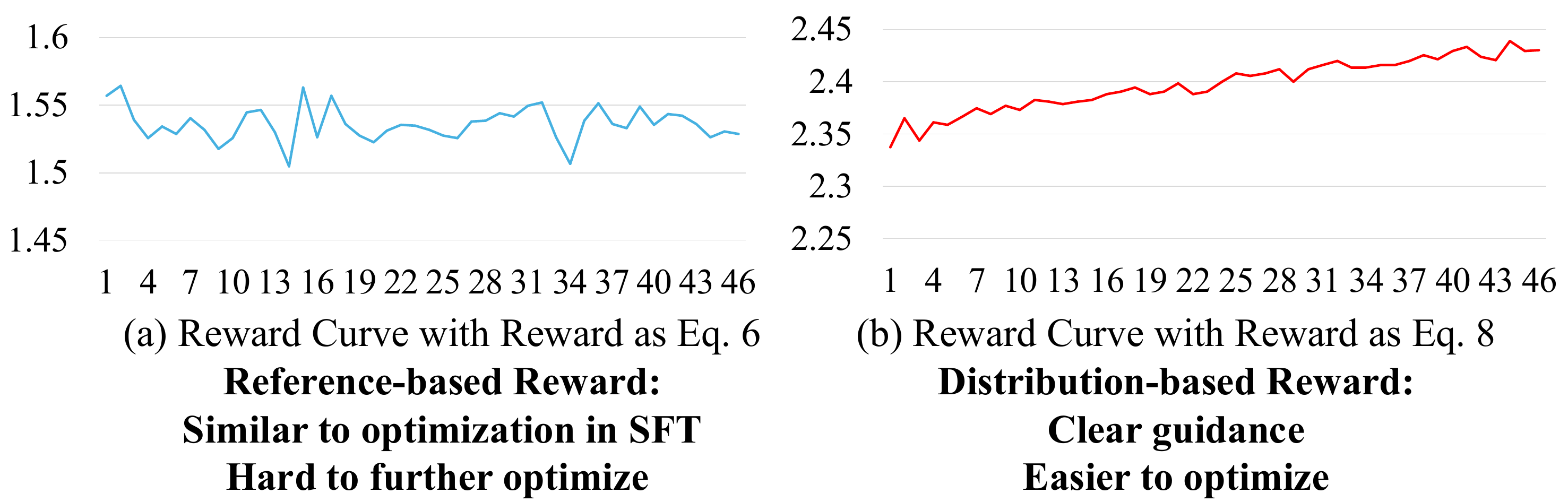}
	\end{center}
    \vspace{-0.2in}
    \caption{Reward curve comparisons using reconstruction error and IQA as the reward function. We observe that the reward guided by IQA increases steadily, whereas the reward based on reconstruction error exhibits noticeable fluctuations.
	}
	\label{fig:reward}
    \vspace{-0.2in}
\end{figure}

\noindent\textbf{Reward with IQA score: overview.}
In this work, we found that the IQA model is a rational choice for formulating the reward function. 
During SFT training, there are always some samples whose outputs from generative diffusion models cannot be well aligned with the ground truth. This occurs because these samples have reached a local optimization point that cannot be further improved. 
With IQA models, RL first help achieve the better distribution alignment between these model outputs and ground truths, helping to escape from the local optimization point by temporarily diverging from the ground-truth-based optimization direction (Fig.~\ref{fig:basic}).
Aligning distribution prompts the following fine-grained alignment with the ground truth, since samples lie within the similar distribution will have smaller Wasserstein distance, making alignment easier than between samples from different distributions. 
This foundation is built based on the optimal transport theory, as
\begin{equation}
\resizebox{0.87\linewidth}{!}{$
    \mathcal{O}(\{\boldsymbol{y}_i\}, \{\boldsymbol{g}_i\}| \boldsymbol{y}_i\in \mathcal{A}, \boldsymbol{g}_i\in \mathcal{A}) \leq \mathcal{O}(\{\boldsymbol{y}_i\}, \{\boldsymbol{g}_i\}| \boldsymbol{y}_i\in \mathcal{A}, \boldsymbol{g}_i\in \mathcal{B})$},
\end{equation}
where $\mathcal{O}$ is the optimal transport value~\cite{villani2008optimal,adrai2023deep,tang2024residual} and $\mathcal{A}$ and $\mathcal{B}$ are two distributions.

\noindent\textbf{Reward with IQA score: distribution alignment.} 
We will introduce the implementation of our RL and describe the distribution alignment phase first. 
Compared with traditional IQA models~\cite{lpips, musiq, ssim, niqe}, MLLM-based IQA models~\cite{coinstruct, qground, depictqa, depictqav2} have shown the better ability to effectively regress accurate quality scores~\cite{you2025teaching}.
We found that restored images with unsatisfactory restoration typically exhibit low IQA scores, as they differ from real images, i.e., the ground truth. These images often contain unrealistic artifacts (e.g., visual degradation or incorrectly generated objects that violate the laws of physics). Fortunately, we discovered that such artifacts can be detected by MLLM-based IQA models (Fig.~\ref{fig:teaser}), as they are trained to assign high scores only to real, high-quality images. Therefore, MLLM-based IQA models can effectively serve as reward functions to distinguish two distributions. Let the MLLM-based IQA score model be denoted as $M$; then, the reward function in Eq.~\ref{eq:policy-gradient} can be
\begin{equation}
    r(\hat{\boldsymbol{x}}_{t-1}, \boldsymbol{c})=M(\hat{\boldsymbol{x}}_{t-1}).
\end{equation}

\subsection{Training Diffusion-based RL with Adaptive Manner for Fine-grained Alignment}
\label{sec:adaptive}

It is important to note that RL with only the IQA reward lacks exact supervision, whereas restoration tasks do have such supervision. 
Therefore, once the network has identified a potential alternative generation path through RL with IQA, it should shift its focus back to optimizing towards the ground truth. In this work, we implement this transition using a difficulty-adaptive attention mechanism combined with SFT, as shown in Fig.~\ref{fig:basic} (b). 

\noindent\textbf{Implementation.} As mentioned above, the role of RL is to explore alternative generation directions for better distribution alignment first. When samples' output distribution has been aligned with the ground truths, further exploration is unnecessary. Therefore, the primary impact of diffusion-based RL is on samples that are far away from the ground truth—i.e., the hard samples with large loss values.

Suppose a batch of data is denoted as ${\boldsymbol{b}_1, ..., \boldsymbol{b}_m}$ with $m$ samples. We first compute their inference results as ${\boldsymbol{y}_1, ..., \boldsymbol{y}_m}$, and denote the corresponding ground truth as ${\boldsymbol{g}_1, ..., \boldsymbol{g}_m}$. Based on this, we calculate the reconstruction differences between ${\boldsymbol{y}_1, ..., \boldsymbol{y}_m}$ and ${\boldsymbol{g}_1, ..., \boldsymbol{g}_m}$, which reflect the difficulty of each sample—more difficult samples exhibit larger reconstruction errors. In this paper, we adopt a strategy of assigning loss weights to train diffusion-based RL in an adaptive manner, where the loss weight is
\begin{equation}
    w_i = \Vert \boldsymbol{y}_i - \boldsymbol{g}_i \Vert / \mathrm{max}(\Vert \boldsymbol{y}_j - \boldsymbol{g}_j \Vert, j\in [1,m]),
    \label{loss_weight}
\end{equation}
where $\mathrm{max}$ is the maximize value computation in this batch.
Loss weights $w_i$ are incorporated into the diffusion-based RL loss as defined in Eq.~\ref{eq:policy-gradient}. In this way, samples that are closer to the distribution of ground truth are assigned smaller RL loss weights, allowing fine-grained reference-based alignment with SFT.

In this way, the final loss function is a combination of the diffusion loss and the diffusion-based RL loss. The diffusion-based RL loss, weighted by the difficulty measure, encourages hard samples to explore alternative optimization directions for distribution alignment. When closer to the admired distribution, diffusion loss is employed for SFT with higher weight.
In summary, for a batch of data ${\boldsymbol{b}_1, ..., \boldsymbol{b}_m}$, the loss term is computed as follows
\begin{equation}
\small
    L_{\boldsymbol{\theta}} = \sum_{i=1:m} (1-w_i)L_{\text{diff}, \boldsymbol{\theta}}(\boldsymbol{b}_i, \boldsymbol{g}_i) + \sum_{i=1:m} w_i \times J(\boldsymbol{\theta}) (\boldsymbol{b}_i) + L_{\text{KL}, \boldsymbol{\theta}},
    \label{eq:loss}
\end{equation}
where $L_{\text{diff}, \boldsymbol{\theta}}$ is the diffusion loss, and $L_{\text{KL}, \boldsymbol{\theta}}$ is optionally added to further stabilize training, ensuring that the model after RL does not deviate significantly from the original reference model~\cite{black2023training}. This is implemented by measuring the difference between $p_{\boldsymbol{\theta}}$ and $p_{\boldsymbol{\theta}_\text{old}}$.

The overall pipeline can be visualized in Figs.~\ref{fig:pipeline} and \ref{fig:pipeline2}.

%% file: sec/4_experiment.tex
\begin{table}[t]
	\centering
	\huge
	\caption{The quantitative comparison between current SOTA methods and their versions with our strategy on LOL-real and LOL-synthetic. ``+Diff.SFT'' means using the selected hard samples for further supervised finetuning on the pretrained model. ``+RL'' denotes applying our RL strategy.}
	\label{comparison1}
	\resizebox{1.0\linewidth}{!}{
		\begin{tabular}{|l|p{2cm}<{\centering}p{2cm}<{\centering}p{2cm}<{\centering}p{2cm}<{\centering}|p{2cm}<{\centering}p{2cm}<{\centering}p{2cm}<{\centering}p{2cm}<{\centering}|}
			\hline
			& \multicolumn{4}{c|}{LOL-real} & \multicolumn{4}{c|}{LOL-synthetic}\\
			\hline
			Methods & PSNR$\uparrow$& SSIM$\uparrow$&LPIPS$\downarrow$&FID$\downarrow$& PSNR$\uparrow$& SSIM$\uparrow$&LPIPS$\downarrow$&FID$\downarrow$\\
			\hline \hline
			DiffBIR & 16.89 &0.717 &0.1139 &88.61 &20.25  &0.752 &0.1004 &40.17  \\
			+Diff.SFT & 17.35 &0.720 &0.0988 &80.66 &20.56 &0.750 & 0.0953&38.29   \\ 
			+Our RL & \textbf{22.11} &\textbf{0.744} & \textbf{0.0846}&\textbf{64.58} &\textbf{21.27} &\textbf{0.755} &\textbf{0.0711} &\textbf{35.47}   \\ \hline
			
			StableSR & 20.39 &0.735 &0.1227 &76.71& 23.42 & 0.784& 0.1173&42.66   \\
			+Diff.SFT  & 21.68 & 0.741&0.1072 &74.79 &24.06 &0.801 &0.0918 & 40.71  \\
			+Our RL  &\textbf{22.47}  & \textbf{0.754}&\textbf{0.0918} &\textbf{71.22} &\textbf{24.87} &\textbf{0.812} & \textbf{0.0901}&\textbf{38.63}   \\  \hline
			PASD  & 20.58 & 0.729& 0.1095 &78.89 & 22.86 &0.780 &0.0935 &38.76  \\
			+Diff.SFT  & 21.43 &0.732 &0.1041 &76.48 &23.17 & 0.791&0.0827 & 37.05  \\ 
			+Our RL  &\textbf{22.46}  &\textbf{0.753} &\textbf{0.0967} &\textbf{73.90} &\textbf{24.02} &\textbf{0.803} & \textbf{0.0806}& \textbf{35.44}  \\   \hline
			XPSR~\cite{qu2024xpsr} &21.15 &0.730 &0.1003 &75.47& 23.04 &0.786& 0.0918& 36.28  \\
			+Diff.SFT & 21.46 &0.735 &0.0980 &74.01 &23.28 &0.792 & 0.0893&34.57\\
			+Our RL&\textbf{22.03} & \textbf{0.746}&  \textbf{0.0939}&  \textbf{71.12}&\textbf{23.90} &\textbf{0.798} &\textbf{0.0841} &\textbf{33.05}\\
			\hline
			TSD-SR~\cite{dong2024tsd}& 21.24& 0.737&0.1026 & 77.83& 23.15&0.769 &0.0954 &38.42\\
			+Diff.SFT &21.82 &0.741 &0.1004 &76.09 &23.46 &0.770 &0.0928 &36.73\\
			+Our RL&\textbf{22.46} &\textbf{0.748} &\textbf{0.0971} & \textbf{73.60}&\textbf{24.07} &\textbf{0.775} & \textbf{0.0890}&\textbf{34.84}\\
			\hline
			RAP~\cite{wang2024rap} &21.79 & 0.741&0.1042 &79.55 & 23.48&0.753 &0.0972 &39.50\\
			+Diff.SFT &21.94 &0.746 &0.1015 &77.93 &23.87 &0.759 &0.0941 &37.82\\
			+Our RL&\textbf{22.50} &\textbf{0.752} &\textbf{0.0968} &\textbf{75.14} &\textbf{24.53} &\textbf{0.764} &\textbf{0.0886} &\textbf{35.17}\\
			\hline
			FaithDiff~\cite{chen2024faithdiff} &22.05 &0.749 &0.0934 &74.07 & 23.92&0.771 &0.0883 &35.61\\
			+Diff.SFT &22.37 &0.755 &0.0902 &73.19 &24.16 &0.778 &0.0861 &34.23\\
			+Our RL&\textbf{23.12} &\textbf{0.763} &\textbf{0.0871} &\textbf{71.68} &\textbf{24.83} & \textbf{0.785}& \textbf{0.0829}&\textbf{32.08}\\
			\hline
			Pixel~\cite{sun2024pixel} & 21.08& 0.724&0.0987 & 78.46&23.36 &0.750 &0.0975 &40.24\\
			+Diff.SFT &21.49 &0.728 &0.0956 &77.04 &23.67 & 0.755&0.0950 &38.43\\
			+Our RL&\textbf{22.17} &\textbf{0.740} & \textbf{0.0905}&\textbf{72.19} &\textbf{24.21} & \textbf{0.762}&\textbf{0.0893} &\textbf{36.59}\\
			\hline
	\end{tabular}}
	\vspace{-0.2in}
\end{table}

\begin{table}[t]
	\centering
	\huge
	\caption{The quantitative comparison between current SOTA methods and their versions with our strategy on SID and SMID. ``+Diff.SFT'' means using the selected hard samples for further supervised finetuning on the pretrained model. ``+RL'' denotes applying our RL strategy.}
	\label{comparison1-1}
	\resizebox{1.0\linewidth}{!}{
		\begin{tabular}{|l|p{2cm}<{\centering}p{2cm}<{\centering}p{2cm}<{\centering}p{2cm}<{\centering}|p{2cm}<{\centering}p{2cm}<{\centering}p{2cm}<{\centering}p{2cm}<{\centering}|}
			\hline
			& \multicolumn{4}{c|}{SID} & \multicolumn{4}{c|}{SMID}\\
			\hline
			Methods & PSNR$\uparrow$& SSIM$\uparrow$&LPIPS$\downarrow$&FID$\downarrow$& PSNR$\uparrow$& SSIM$\uparrow$&LPIPS$\downarrow$&FID$\downarrow$\\
			\hline \hline
			DiffBIR & 17.85 &0.604&0.2178 &90.62 & 22.47 &0.763 & 0.1836& 88.21\\
			+Diff.SFT  &20.75  &0.619 &0.1882 &87.68 &23.81 & 0.778&0.1798 &83.09  \\ 
			+Our RL & \textbf{22.48} &\textbf{0.634} &\textbf{0.1733} &\textbf{84.64} &\textbf{24.56} &\textbf{0.781} &\textbf{0.1672} & \textbf{80.47} \\ \hline
			
			StableSR & 20.27 &0.620 &0.2074 &87.39 & 24.08 &0.773 &0.1798 &85.75 \\
			+Diff.SFT   &  21.38&0.633&0.1895&85.76 &24.63 &0.780 &0.1727 &83.82  \\
			+Our RL  &  \textbf{22.04}&\textbf{0.662}&\textbf{0.1758}&\textbf{83.87} &\textbf{25.37} &\textbf{0.790} &\textbf{0.1601} &\textbf{80.94}  \\  \hline
			PASD & 20.62 &0.674& 0.1958 &81.83 &24.78  &0.780 &0.1856 &83.14 \\
			+Diff.SFT   & 20.90&0.686&0.1829&78.96 &24.97 & 0.786& 0.1809& 81.30 \\ 
			+Our RL  &  \textbf{21.91}&\textbf{0.703}&\textbf{0.1800}&\textbf{76.75}& \textbf{25.37}&\textbf{0.792} &\textbf{0.1778} &\textbf{78.95}  \\   \hline
	\end{tabular}}
	\vspace{-0.2in}
\end{table}

\begin{table*}[t]
\begin{center}
\caption{The comparison for image deraining results. }
\label{table:deraining}
\vspace{-0.1in}
\resizebox{1.0\linewidth}{!}{
\begin{tabular}{|l| c c| c c| c c|c c| c c| c c|}
\hline
 \textbf{Method} & PSNR~$\textcolor{black}{\uparrow}$ & SSIM~$\textcolor{black}{\uparrow}$ & PSNR~$\textcolor{black}{\uparrow}$ & SSIM~$\textcolor{black}{\uparrow}$ & PSNR~$\textcolor{black}{\uparrow}$ & SSIM~$\textcolor{black}{\uparrow}$ & PSNR~$\textcolor{black}{\uparrow}$ & SSIM~$\textcolor{black}{\uparrow}$ & PSNR~$\textcolor{black}{\uparrow}$ & SSIM~$\textcolor{black}{\uparrow}$ & PSNR~$\textcolor{black}{\uparrow}$ & SSIM~$\textcolor{black}{\uparrow}$ \\
\hline \hline
  & \multicolumn{2}{c|}{\textbf{Test100}}&\multicolumn{2}{c|}{\textbf{Rain100H}}&\multicolumn{2}{c|}{\textbf{Rain100L}}& \multicolumn{2}{c|}{\textbf{Test2800}}&\multicolumn{2}{c|}{\textbf{Test1200}}&\multicolumn{2}{c|}{\textbf{Mean Value}}\\ \hline
DiffBIR & {24.50} & {0.707} & {24.84} & {0.702} & {30.83} & {0.765} &27.75 & 0.734 & {27.26} &{0.721} & {27.04} & {0.726} \\
DiffBIR+Ours  & \textbf{25.78} &\textbf{0.739}   &\textbf{25.62} &\textbf{0.724} &\textbf{31.82} &\textbf{0.781} &\textbf{28.99}  &\textbf{0.750} &\textbf{28.54}  &\textbf{0.738}  &\textbf{28.15} &\textbf{0.746} \\ \hline
PASD& {25.30} & {0.718} & {25.59} & {0.723} & {31.17} & {0.776}& {28.79} & {0.748} & {28.54} & {0.737} & {27.88} & {0.740}   \\
PASD+Ours  &\textbf{26.87}  &\textbf{0.740}   &\textbf{26.46} &\textbf{0.745} &\textbf{32.04} &\textbf{0.792} &\textbf{29.37}  &\textbf{0.763} &\textbf{29.25}  &\textbf{0.750}  &\textbf{28.80} & \textbf{0.758}\\
\hline
\end{tabular}}
\vspace{-0.2in}
\end{center}
\end{table*}

\begin{table}[!t]
\begin{center}
\caption{Single-image motion deblurring results. }
\label{tab:motion}
\resizebox{1.0\linewidth}{!}{
\begin{tabular}{|l| p{0.8cm}<{\centering}p{0.8cm}<{\centering} | p{0.8cm}<{\centering}p{0.8cm}<{\centering} | p{0.8cm}<{\centering}p{0.8cm}<{\centering} | p{0.8cm}<{\centering}p{0.8cm}<{\centering} |}
\hline
 \multirow{2}{1cm}{\textbf{Method}} & \multicolumn{2}{c|}{\textbf{GoPro}} & \multicolumn{2}{c|}{\textbf{HIDE}} & \multicolumn{2}{c|}{\textbf{RealBlur-R}} & \multicolumn{2}{c|}{\textbf{\textbf{RealBlur-J}}} \\
  & PSNR$\textcolor{black}{\uparrow}$ & {SSIM}$\textcolor{black}{\uparrow}$ & PSNR$\textcolor{black}{\uparrow}$ & {SSIM}$\textcolor{black}{\uparrow}$ & PSNR$\textcolor{black}{\uparrow}$ & {SSIM}$\textcolor{black}{\uparrow}$ & PSNR$\textcolor{black}{\uparrow}$ &{SSIM}$\textcolor{black}{\uparrow}$\\
\hline \hline
DiffBIR & {27.99} & {0.862} &	{26.88} &{{0.844}} & {31.87} &  {0.853} & 24.53& {0.772} \\
+Ours &\textbf{29.76} &\textbf{0.887} &\textbf{28.43}  &\textbf{0.862} &\textbf{33.09}  &\textbf{0.872} &\textbf{25.67} &\textbf{0.790}\\ \hline
PASD & {28.67} & {{0.881}} & {27.41} & {{0.858}} & {32.64} & {{0.869}} & {25.36} & {{0.793}}\\
+Ours &\textbf{30.02} &\textbf{0.893} &\textbf{28.49}  &\textbf{0.870} &\textbf{33.87}  &\textbf{0.876} &\textbf{26.80} &\textbf{0.802}\\
\hline
\end{tabular}}
\vspace{-0.2in}
\end{center}
\end{table}

\begin{table}[!t]
\begin{center}
\caption{Defocus deblurring comparisons on the DPDD testset (containing 37 indoor and 39 outdoor scenes). \textbf{S:} single-image defocus deblurring. \textbf{D:} dual-pixel defocus deblurring. \textbf{I. S.} and \textbf{O. S.} mean ``Indoor Scenes" and ``Outdoor Scenes", respectively.
}
\label{table:dpdeblurring}
\vspace{-0.1in}
\resizebox{1.0\linewidth}{!}{
\begin{tabular}{|l | p{0.8cm}<{\centering}p{0.8cm}<{\centering}  | p{0.8cm}<{\centering}p{0.8cm}<{\centering}  | p{0.8cm}<{\centering}p{0.8cm}<{\centering} |p{0.8cm}<{\centering}p{0.8cm}<{\centering} |}
\hline
   \multirow{2}{1cm}{\textbf{Method}} & \multicolumn{2}{c|}{\textbf{I. S.} (S)} & \multicolumn{2}{c|}{\textbf{O. S.} (S)} & \multicolumn{2}{c|}{\textbf{I. S.} (D)} & \multicolumn{2}{c|}{\textbf{O. S.} (D)}  \\
\cline{2-9}
    & PSNR$\textcolor{black}{\uparrow}$ & SSIM$\textcolor{black}{\uparrow}$ & PSNR$\textcolor{black}{\uparrow}$ & SSIM$\textcolor{black}{\uparrow}$ & PSNR$\textcolor{black}{\uparrow}$ & SSIM$\textcolor{black}{\uparrow}$& PSNR$\textcolor{black}{\uparrow}$ & SSIM$\textcolor{black}{\uparrow}$  \\
\hline \hline
{DiffBIR}& {25.34}  & {0.808}  & {20.07}  & {0.663}   & {25.91}  & {0.835}   & {20.67}  & {0.684}     \\
+Ours &\textbf{26.28} &\textbf{0.829}  & \textbf{21.62} & \textbf{0.679} &\textbf{27.20} &\textbf{0.844}  &\textbf{21.05} &\textbf{0.697}   \\ \hline
PASD &27.36 & 0.844  & 20.75 &0.702 & 27.02& 0.863 & 21.90& 0.708\\
+Ours &\textbf{27.97} &\textbf{0.866}  &\textbf{21.73}  &\textbf{0.721}  &\textbf{27.74} & \textbf{0.872} &\textbf{22.43} &\textbf{0.720}   \\
\hline
\end{tabular}}
\vspace{-0.2in}
\end{center}
\end{table}

\begin{figure*}[t]
	\centering

	\begin{subfigure}[c]{0.12\textwidth}
		\centering
		\includegraphics[width=0.85in,height=0.85in]{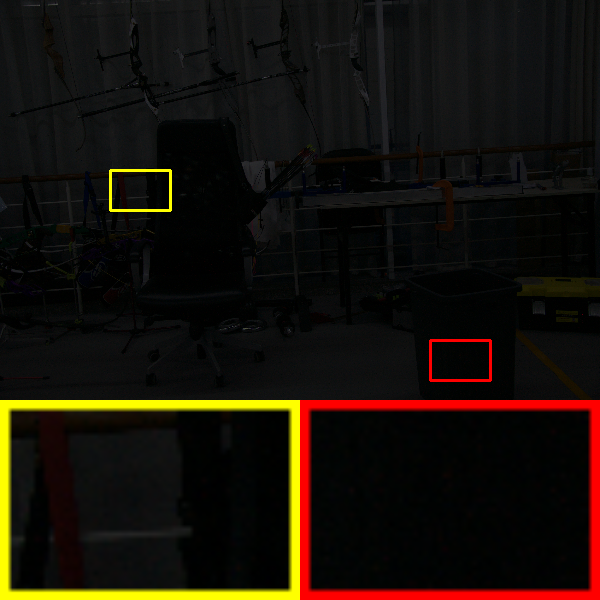}
		\vspace{-1.5em}
		\caption*{LOL-real}
	\end{subfigure}
	\begin{subfigure}[c]{0.12\textwidth}
		\centering
		\includegraphics[width=0.85in,height=0.85in]{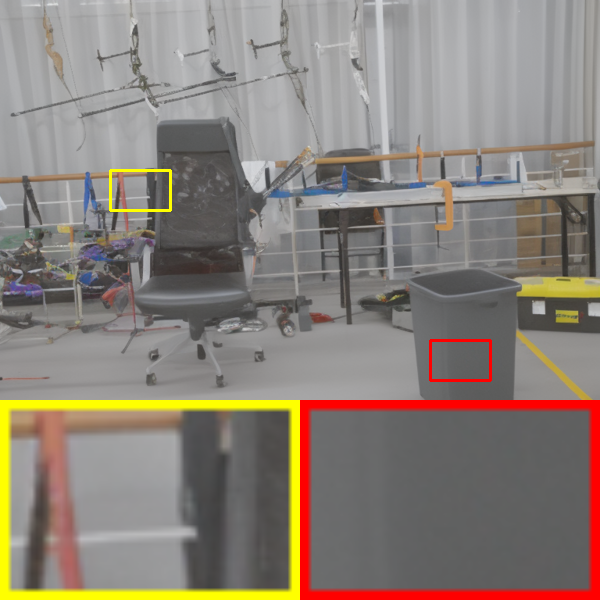}
		\vspace{-1.5em}
		\caption*{DiffBIR}
	\end{subfigure}
	\begin{subfigure}[c]{0.12\textwidth}
		\centering
		\includegraphics[width=0.85in,height=0.85in]{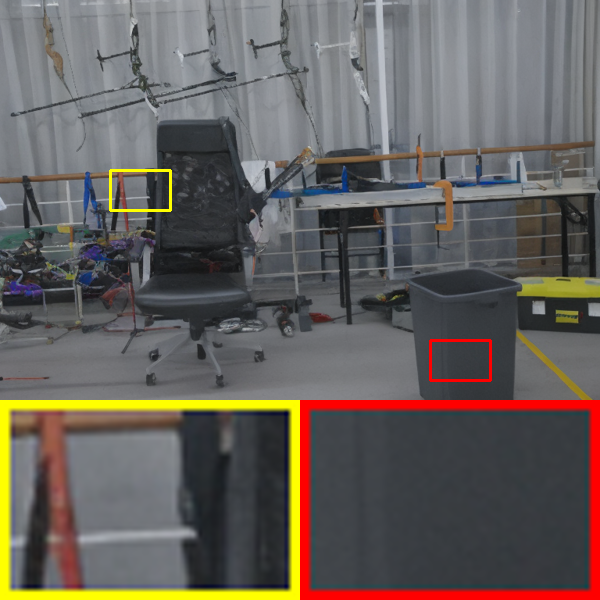}
		\vspace{-1.5em}
		\caption*{+Ours}
	\end{subfigure}
	\begin{subfigure}[c]{0.12\textwidth}
		\centering
		\includegraphics[width=0.85in,height=0.85in]{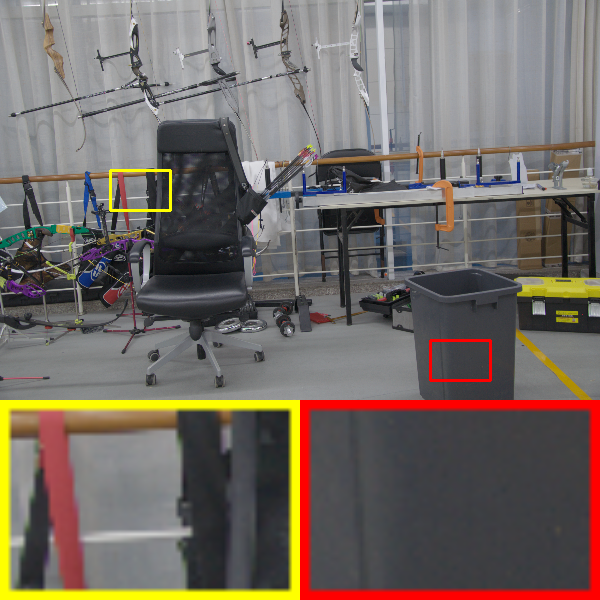}
		\vspace{-1.5em}
		\caption*{GT}
	\end{subfigure}
	\begin{subfigure}[c]{0.12\textwidth}
		\centering
		\includegraphics[width=0.85in,height=0.85in]{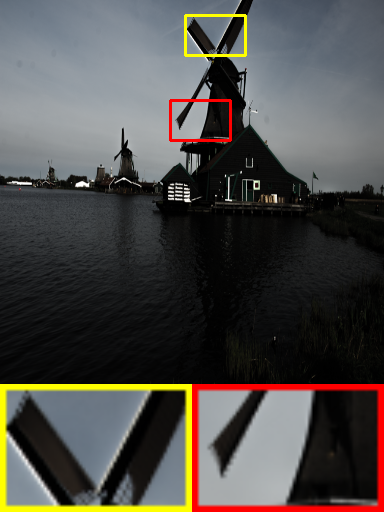}
		\vspace{-1.5em}
		\caption*{LOL-syn.}
	\end{subfigure}
	\begin{subfigure}[c]{0.12\textwidth}
		\centering
		\includegraphics[width=0.85in,height=0.85in]{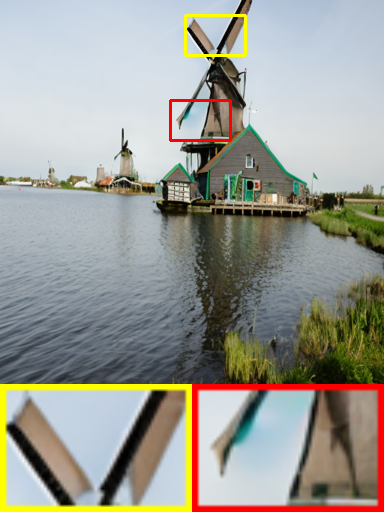}
		\vspace{-1.5em}
		\caption*{DiffBIR}
	\end{subfigure}
	\begin{subfigure}[c]{0.12\textwidth}
		\centering
		\includegraphics[width=0.85in,height=0.85in]{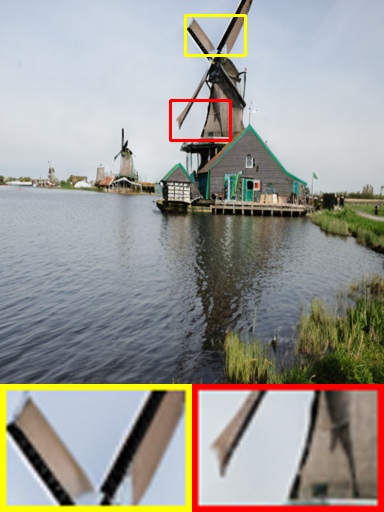}
		\vspace{-1.5em}
		\caption*{+Ours}
	\end{subfigure}
	\begin{subfigure}[c]{0.12\textwidth}
		\centering
		\includegraphics[width=0.85in,height=0.85in]{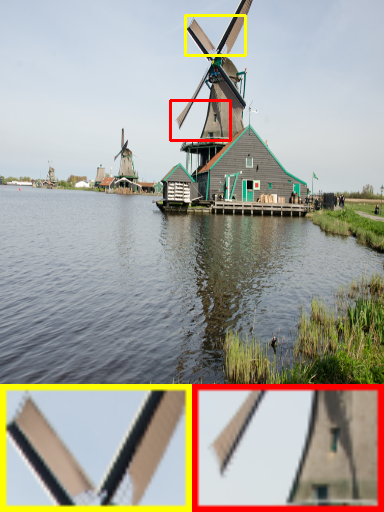}
		\vspace{-1.5em}
		\caption*{GT}
	\end{subfigure}

    \centering
	\begin{subfigure}[c]{0.12\textwidth}
		\centering
		\includegraphics[width=0.85in]{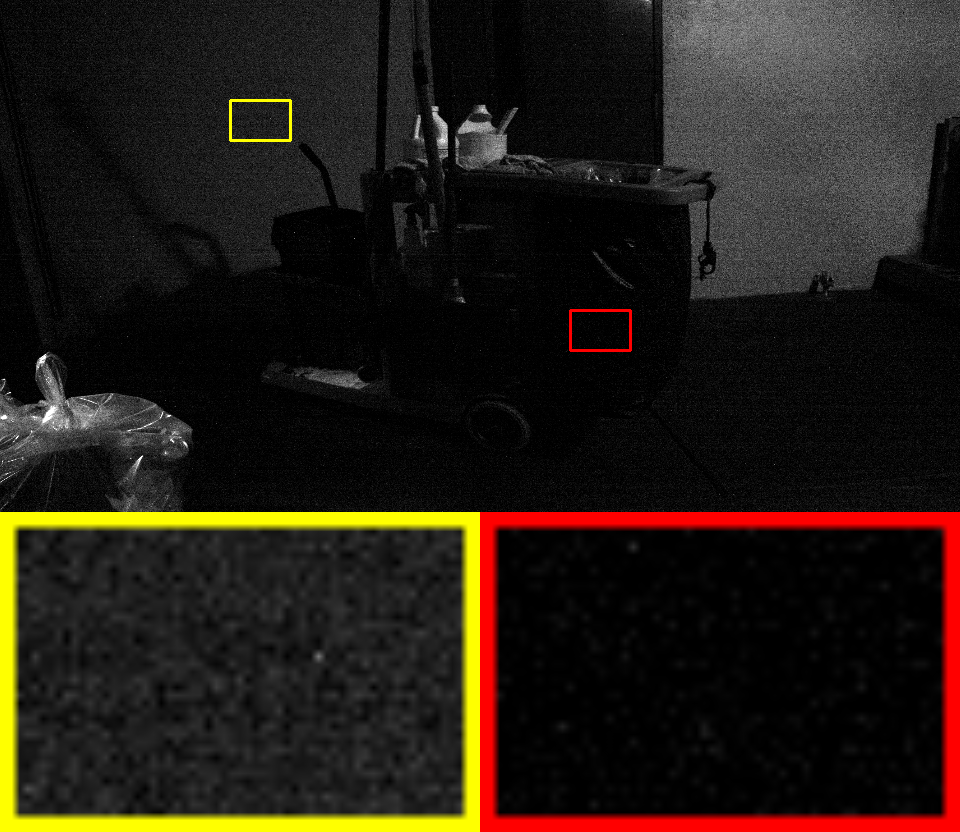}
		\vspace{-1.5em}
		\caption*{SID}
	\end{subfigure}
	\begin{subfigure}[c]{0.12\textwidth}
		\centering
		\includegraphics[width=0.85in]{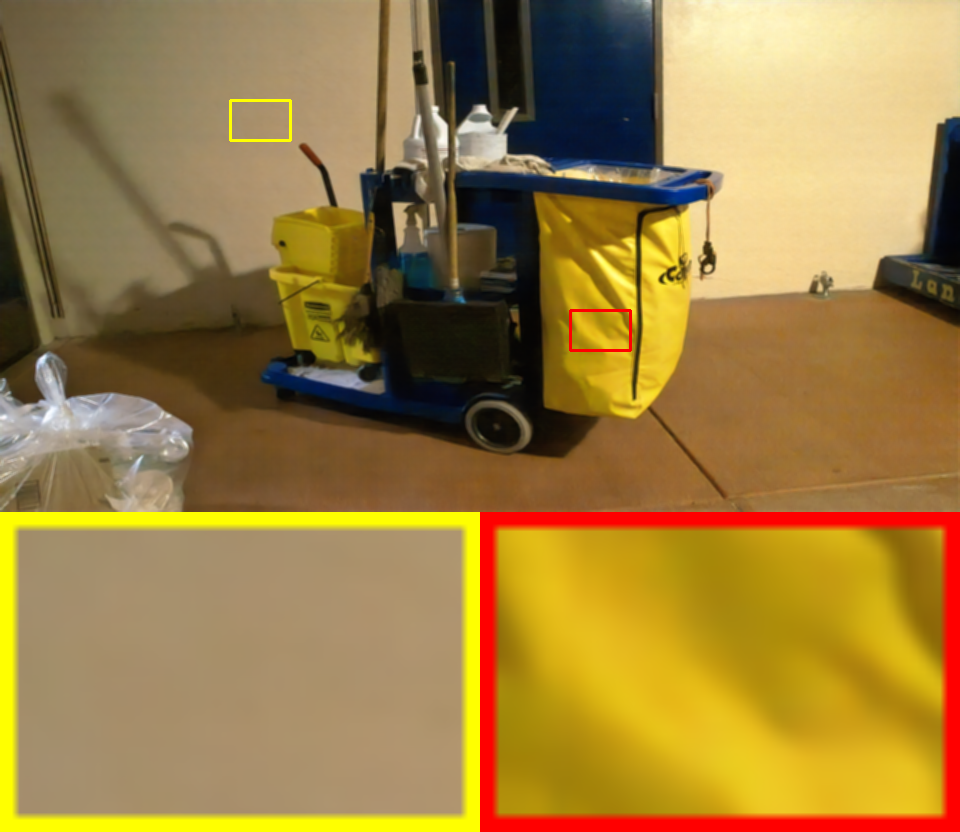}
		\vspace{-1.5em}
		\caption*{StableSR}
	\end{subfigure}
	\begin{subfigure}[c]{0.12\textwidth}
		\centering
		\includegraphics[width=0.85in]{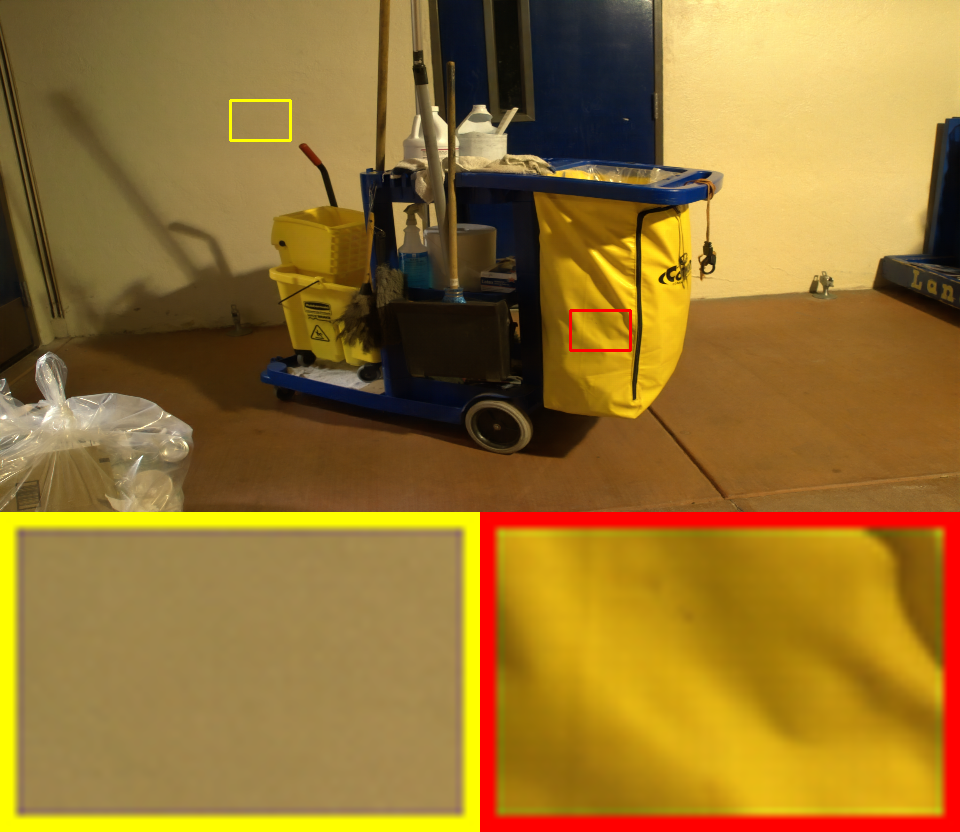}
		\vspace{-1.5em}
		\caption*{+Ours}
	\end{subfigure}
	\begin{subfigure}[c]{0.12\textwidth}
		\centering
		\includegraphics[width=0.85in]{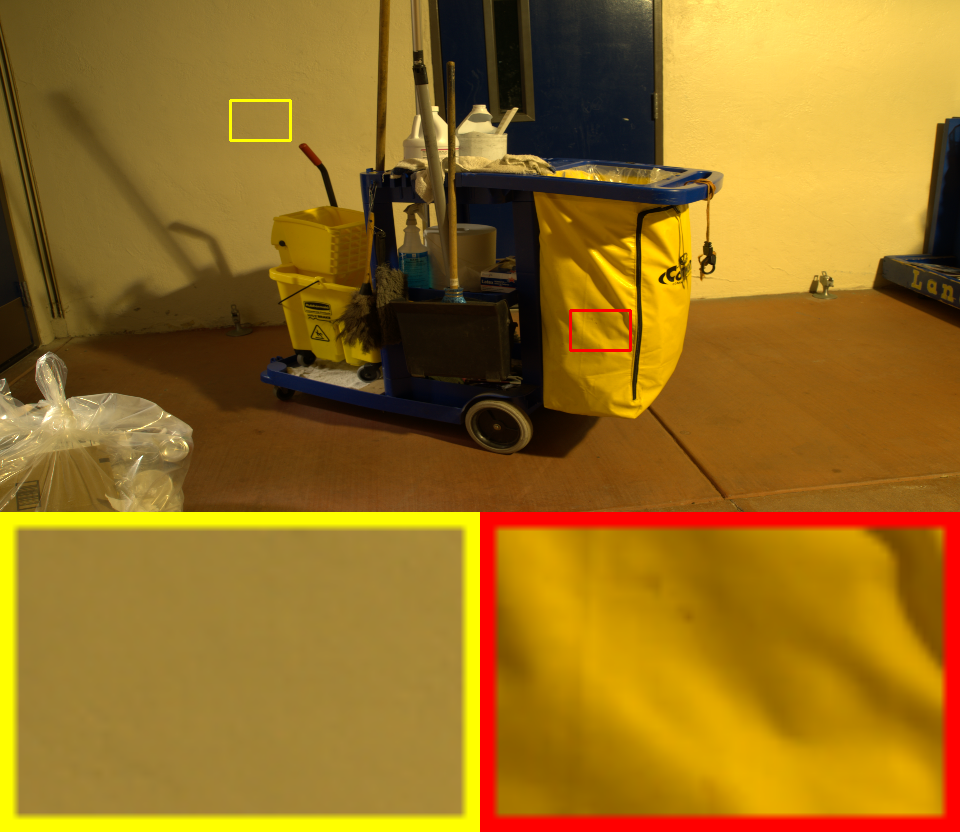}
		\vspace{-1.5em}
		\caption*{GT}
	\end{subfigure}
	\begin{subfigure}[c]{0.12\textwidth}
		\centering
		\includegraphics[width=0.85in]{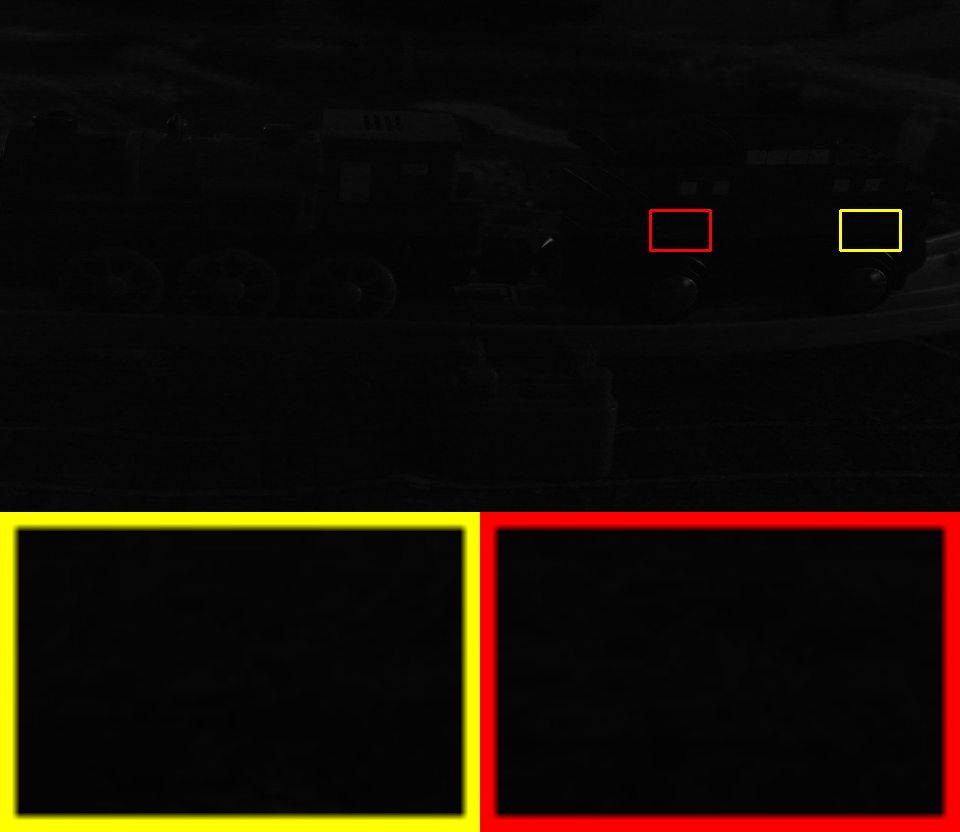}
		\vspace{-1.5em}
		\caption*{SMID}
	\end{subfigure}
	\begin{subfigure}[c]{0.12\textwidth}
		\centering
		\includegraphics[width=0.85in]{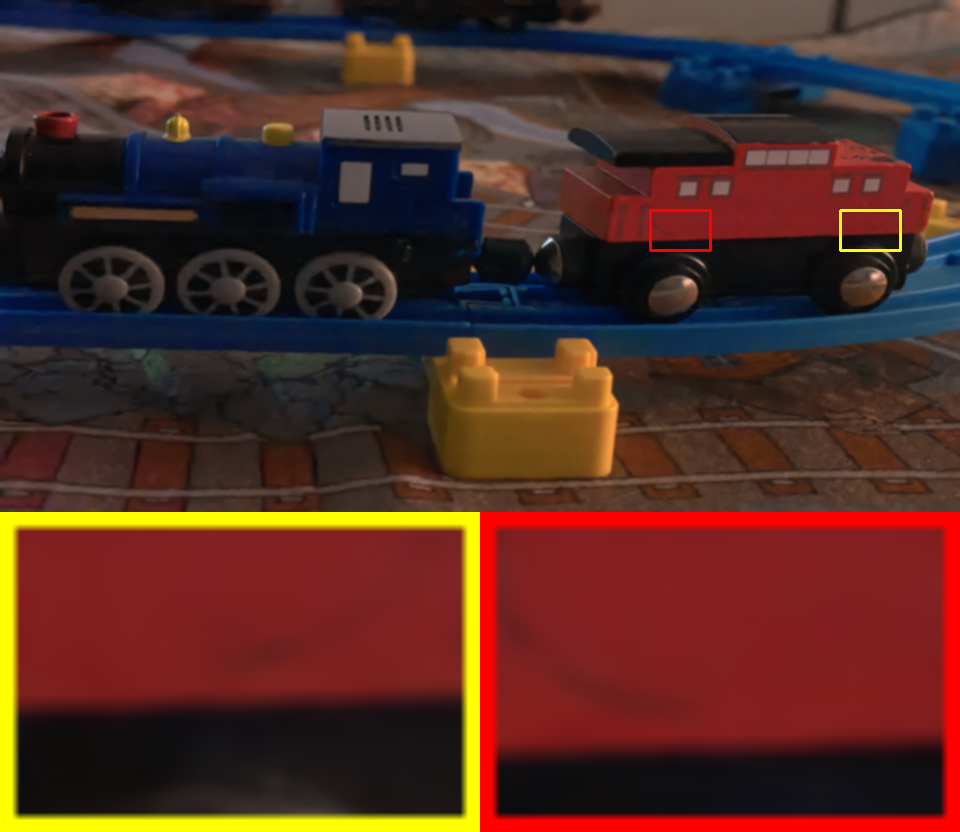}
		\vspace{-1.5em}
		\caption*{PASD}
	\end{subfigure}
	\begin{subfigure}[c]{0.12\textwidth}
		\centering
		\includegraphics[width=0.85in]{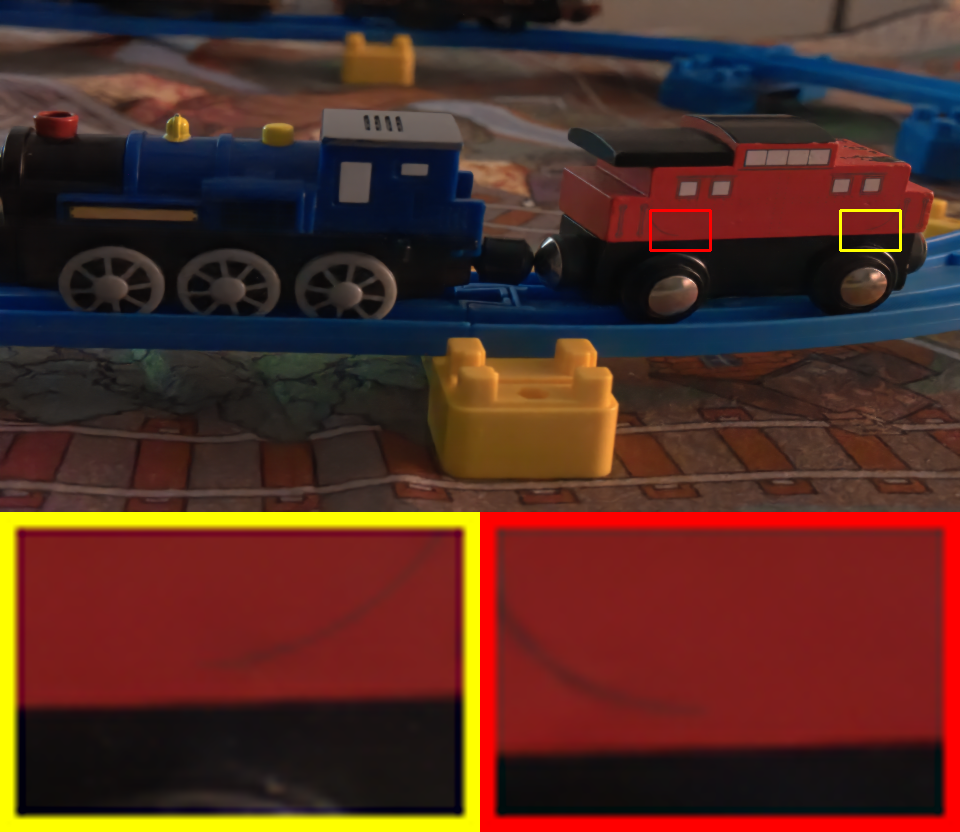}
		\vspace{-1.5em}
		\caption*{+Ours}
	\end{subfigure}
	\begin{subfigure}[c]{0.12\textwidth}
		\centering
		\includegraphics[width=0.85in]{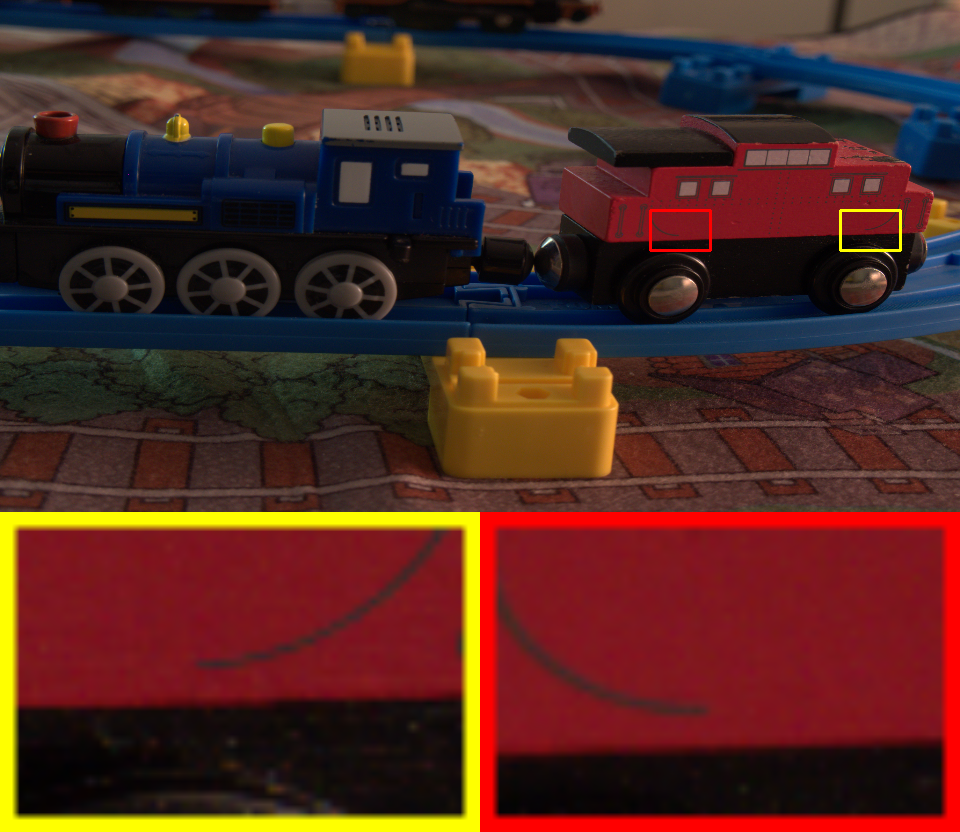}
		\vspace{-1.5em}
		\caption*{GT}
	\end{subfigure}

	\begin{subfigure}[c]{0.12\textwidth}
		\centering
		\includegraphics[width=0.85in,height=0.64in]{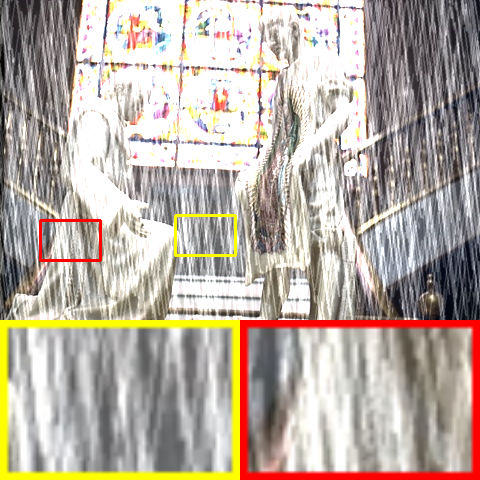}
		\vspace{-1.5em}
		\caption*{Derain}
	\end{subfigure}
	\begin{subfigure}[c]{0.12\textwidth}
		\centering
		\includegraphics[width=0.85in,height=0.64in]{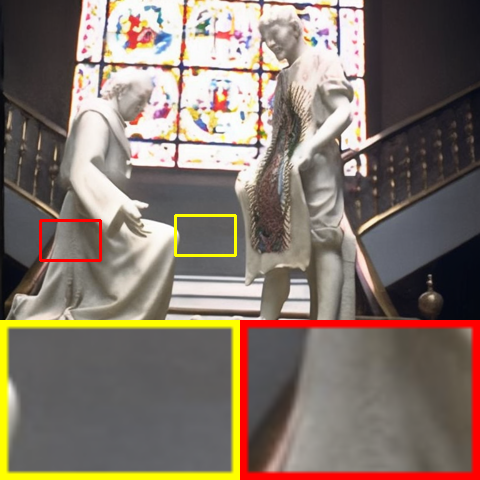}
		\vspace{-1.5em}
		\caption*{PASD}
	\end{subfigure}
	\begin{subfigure}[c]{0.12\textwidth}
		\centering
		\includegraphics[width=0.85in,height=0.64in]{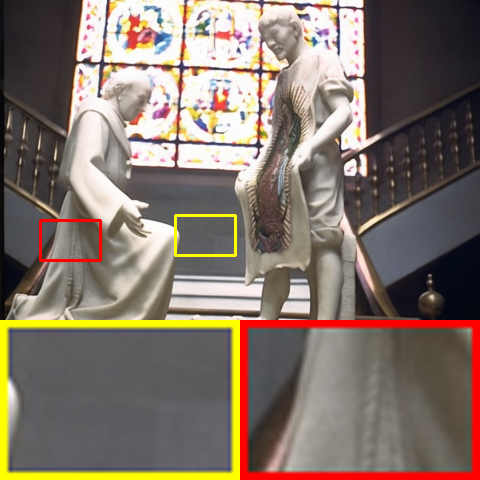}
		\vspace{-1.5em}
		\caption*{+Ours}
	\end{subfigure}
	\begin{subfigure}[c]{0.12\textwidth}
		\centering
		\includegraphics[width=0.85in,height=0.64in]{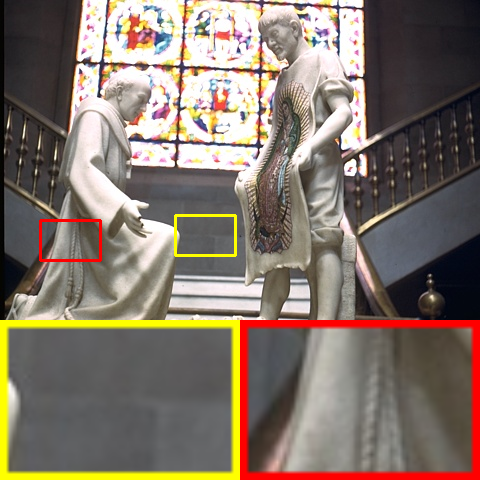}
		\vspace{-1.5em}
		\caption*{GT}
	\end{subfigure}
	\begin{subfigure}[c]{0.12\textwidth}
		\centering
		\includegraphics[width=0.85in,height=0.64in]{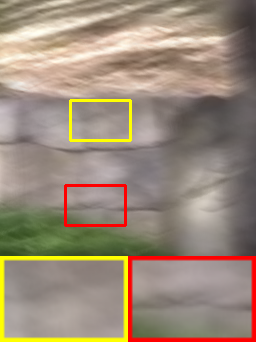}
		\vspace{-1.5em}
		\caption*{M.Deblur}
	\end{subfigure}
	\begin{subfigure}[c]{0.12\textwidth}
		\centering
		\includegraphics[width=0.85in,height=0.64in]{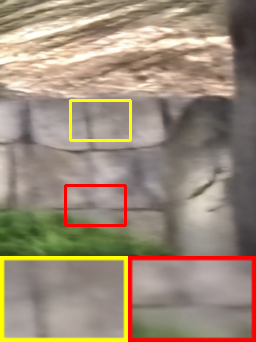}
		\vspace{-1.5em}
		\caption*{PASD}
	\end{subfigure}
	\begin{subfigure}[c]{0.12\textwidth}
		\centering
		\includegraphics[width=0.85in,height=0.64in]{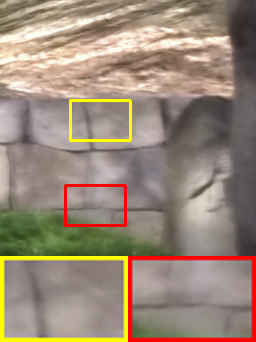}
		\vspace{-1.5em}
		\caption*{+Ours}
	\end{subfigure}
	\begin{subfigure}[c]{0.12\textwidth}
		\centering
		\includegraphics[width=0.85in,height=0.64in]{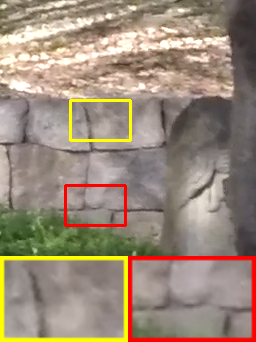}
		\vspace{-1.5em}
		\caption*{GT}
	\end{subfigure}

    \centering
	\begin{subfigure}[c]{0.12\textwidth}
		\centering
		\includegraphics[width=0.85in]{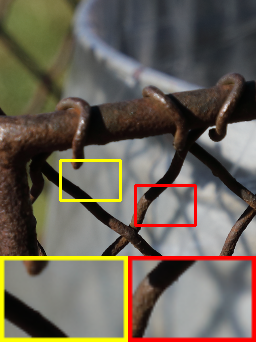}
		\vspace{-1.5em}
		\caption*{D.Deblur(S)}
	\end{subfigure}
	\begin{subfigure}[c]{0.12\textwidth}
		\centering
		\includegraphics[width=0.85in]{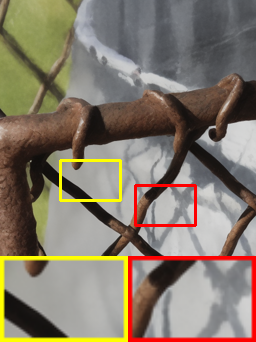}
		\vspace{-1.5em}
		\caption*{PASD}
	\end{subfigure}
	\begin{subfigure}[c]{0.12\textwidth}
		\centering
		\includegraphics[width=0.85in]{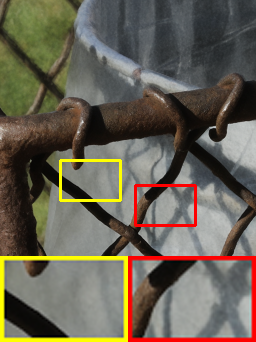}
		\vspace{-1.5em}
		\caption*{+Ours}
	\end{subfigure}
	\begin{subfigure}[c]{0.12\textwidth}
		\centering
		\includegraphics[width=0.85in]{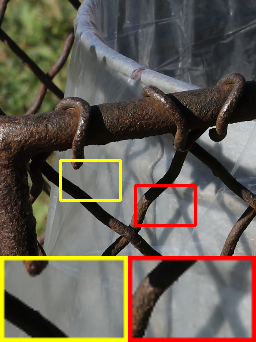}
		\vspace{-1.5em}
		\caption*{GT}
	\end{subfigure}
	\begin{subfigure}[c]{0.12\textwidth}
		\centering
		\includegraphics[width=0.85in]{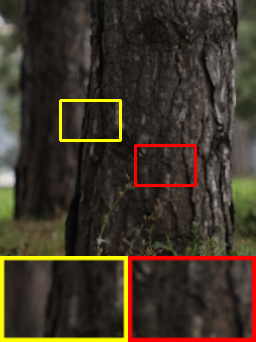}
		\vspace{-1.5em}
		\caption*{D.Deblur(D)}
	\end{subfigure}
	\begin{subfigure}[c]{0.12\textwidth}
		\centering
		\includegraphics[width=0.85in]{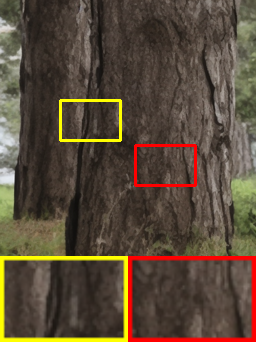}
		\vspace{-1.5em}
		\caption*{PASD}
	\end{subfigure}
	\begin{subfigure}[c]{0.12\textwidth}
		\centering
		\includegraphics[width=0.85in]{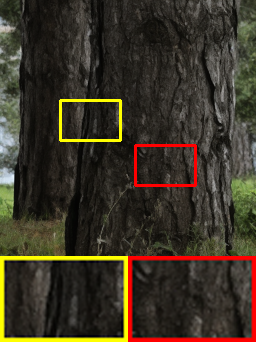}
		\vspace{-1.5em}
		\caption*{+Ours}
	\end{subfigure}
	\begin{subfigure}[c]{0.12\textwidth}
		\centering
		\includegraphics[width=0.85in]{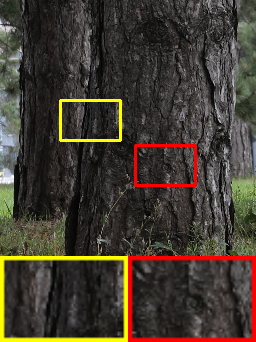}
		\vspace{-1.5em}
		\caption*{GT}
	\end{subfigure}

	\vspace{-0.1in}
\caption{Visual comparisons on different datasets with various network structures. ``LOL-syn.'': LOL-synthetic, ``M.Deblur'': motion deblurring, ``D.Deblur'': defocus deblurring.}
	\label{fig:cmp}
    \vspace{-0.2in}
\end{figure*}

\section{Experiments}
\label{sec:exp}

\subsection{Datasets}
We conduct experiments on several representative image restoration tasks, including low-light image enhancement, deraining, motion and defocus deblurring.
For low-light image enhancement, we evaluate our framework on LOL-real, LOL-synthetic~\cite{yang2021sparse}, SID~\cite{chen2018learning}, and SMID~\cite{chen2019seeing}. 
The experiments are conducted in the sRGB domain.
For deraining, we use the Rain13K~\cite{zamir2022restormer} dataset for training and evaluate on five benchmark datasets: Rain100H~\cite{yang2017deep}, Rain100L~\cite{yang2017deep}, Test100~\cite{zhang2019image}, Test1200~\cite{zhang2018density}, and Test2800~\cite{fu2017removing}.
For single-image motion deblurring, we train on the GoPro~\cite{gopro2017} dataset and evaluate on both synthetic (GoPro~\cite{gopro2017}, HIDE~\cite{shen2019human}) and real-world datasets (RealBlur-R~\cite{rim_2020_realblur}, RealBlur-J~\cite{rim_2020_realblur}).
For defocus deblurring, we use the DPDD~\cite{abdullah2020dpdd} dataset for training and evaluate on EBDB~\cite{karaali2017edge_EBDB} and JNB~\cite{shi2015just_jnb}.

\subsection{Implementation Details}

Traditionally, diffusion-based methods rely on a pre-trained restoration model to extract clean conditional inputs from degraded images. In our experiments, we adopt the SNR-aware network~\cite{xu2022snr} as the pre-trained restoration model for the low-light image enhancement task. For other tasks, such as image deraining and deblurring, we employ Restormer~\cite{zamir2022restormer} as the restoration backbone. Our baselines include several representative diffusion-based methods: StableSR~\cite{wang2024exploiting}, PASD~\cite{yang2024pixel}, and DiffBIR~\cite{lin2024diffbir}. In their original implementations, these methods typically use a text-to-image Stable Diffusion model as the pre-trained diffusion backbone.

\subsection{Comparisons}

To evaluate the effectiveness of our RL strategy in enhancing existing diffusion-based methods, we apply our RL approach to these methods and train them for a limited number of iterations—significantly fewer than those used in the SFT stage. We then observe the resulting performance changes.
In addition to comparing the performance of these diffusion-based methods before and after applying RL, we also evaluate their results when trained for the same number of iterations using the SFT strategy. Unlike the standard SFT stage, this process focuses on fine-tuning the model specifically on difficult samples that exhibit large discrepancies from the ground truth. This is achieved through a weighting scheme defined in Eq.~\ref{loss_weight}.
This forms a meaningful baseline, as the only difference from our method is the absence of the RL process. Results obtained using our RL strategy are denoted as ``+RL'', while those further fine-tuned using the difficulty-aware SFT are denoted as ``+Diff.SFT''.

As shown in Tables~\ref{comparison1} and \ref{comparison1-1}, the superiority of ``+Diff.SFT'' over the baseline highlights the importance of targeted alignment for hard samples. Moreover, the advantage of ``+RL'' over ``+Diff.SFT'' demonstrates that our dynamic RL approach, when combined with SFT, outperforms standalone SFT. This confirms the effectiveness of our RL strategy, which leverages IQA to discover new generation pathways for more effective distribution alignment.

Moreover, experimental results on other tasks are presented in Tables~\ref{table:deraining}, \ref{tab:motion}, and \ref{table:dpdeblurring}, where the advantages of using RL over the baseline are consistently demonstrated. Nontrivial improvements are observed across these tasks, confirming the effectiveness of our approach.
Furthermore, visual comparisons in Fig.~\ref{fig:cmp} illustrate the impact of RL in enhancing both the visual quality and perceptual fidelity.

\begin{table}[t]
	\centering
	\huge
    \caption{The ablation study results. The experiments are conducted with DiffBIR and SNR-aware network as the pretrained diffusion and restoration models.}
    \vspace{-0.1in}
	\label{comparison1-abla}
    \resizebox{1.0\linewidth}{!}{
		\begin{tabular}{|l|p{1.8cm}<{\centering}p{1.8cm}<{\centering}p{1.8cm}<{\centering}p{1.8cm}<{\centering}|p{1.8cm}<{\centering}p{1.8cm}<{\centering}p{1.8cm}<{\centering}p{1.8cm}<{\centering}|}
            \hline
			& \multicolumn{4}{c|}{LOL-real} & \multicolumn{4}{c|}{LOL-synthetic}\\
			\hline
            Methods & PSNR$\uparrow$& SSIM$\uparrow$&LPIPS$\downarrow$&FID$\downarrow$& PSNR$\uparrow$& SSIM$\uparrow$&LPIPS$\downarrow$&FID$\downarrow$\\
			\hline \hline
            Base & 16.89 &0.717 &0.1139 &88.61 &20.25  &0.752 &0.1004 &40.17  \\
            Base +Diff.SFT & 17.35 &0.720 &0.0988 &80.66 &20.56 &0.750 & 0.0953&38.29  \\ 
            with Rec. & 18.54 &0.728  &0.1186 &81.15 &20.88  &0.753 &0.0912 &38.49 \\
            w/o $w_i$ &20.89  &0.727  &0.0944 & 70.73& 20.96 &0.749 &0.0844 &37.22 \\
            with $x_{t-1}$ & 21.46 &0.739  &0.0910 &68.20 & 20.83 &0.750 &0.0835 &37.64 \\
            Reward $x_0$ & 21.73 &0.724 &0.0906 &69.22 &20.72  & 0.735&0.0807 &38.28 \\
            Norm. from track & 21.82 &0.740  & 0.0895& 69.71 & 20.97& 0.744&0.0923 &38.70 \\
            with Q-align & 22.04 & 0.732 &0.0858 &66.87 & 21.08 &0.751 & 0.0816&36.95 \\
            with CLIP-IQA &21.51 &0.725  &0.0923 & 70.76& 20.95 &0.746 &0.0900 &37.01 \\
            \hline
            with Iter. RL &\textbf{23.37}  & \textbf{0.761} &\textbf{0.0810} &\textbf{61.94} & \textbf{22.04} & \textbf{0.762}&\textbf{0.0695} &\textbf{33.86} \\
			Original & \textbf{22.11} &\textbf{0.744} & \textbf{0.0846}&\textbf{64.58} &\textbf{21.27} &\textbf{0.755} &\textbf{0.0711} &\textbf{35.47}  \\
            \hline
	\end{tabular}}
    \vspace{-0.2in}
\end{table}

\subsection{Ablation Study}

In this section, we present several ablation studies to analyze the impact of our proposed different strategies. Here, we select the low-light image enhancement as the example.

\noindent\textbf{The comparison between using reconstruction error and IQA as the reward model.}
As mentioned above, using IQA as the reward model yields superior results compared to using reconstruction error. In this ablation setting, we report the exact results and refer to the reconstruction-based variant as ``with Rec.''. As shown in Table~\ref{comparison1-abla}, the results clearly support the superiority of IQA as reward.
The visual comparison has been placed in Fig.~\ref{fig:reward-visual}.

\noindent\textbf{The effects of conducting RL with adaptive weights.}
One of the main contributions of this paper is the proposed method that assigns different weights to samples based on their difficulty. To evaluate its impact, we conduct an ablation study by removing this weighting mechanism, i.e., excluding $w_i$ as defined Eq.~\ref{loss_weight}. This setting is referred to as ``w/o $w_i$''. The comparison between ``w/o $w_i$'' and ``Original'' in Table~\ref{comparison1-abla} supports the rationale behind our strategy: first applying RL for distribution-level alignment, followed by SFT for fine-grained alignment.

\noindent\textbf{The effects of conducting RL with denoised direction.}
As we have mentioned, better performance can be achieved by optimizing the policy modeling through the use of the more clean $\hat{\boldsymbol{x}}_{t-1}$ instead of $\boldsymbol{x}_{t-1}$.
In this ablation study, we adopt $\boldsymbol{x}_{t-1}$ for experiments instead, and refer to this configuration as ``with $\boldsymbol{x}_{t-1}$''.
As shown in Table~\ref{comparison1-abla}, this variant yields inferior results compared to the original setting, validating the effectiveness of using $\hat{\boldsymbol{x}}_{t-1}$ instead of $\boldsymbol{x}_{t-1}$ for policy modeling.

\noindent\textbf{The effects of conducting RL with rewards from multiple time steps.}
As we have mentioned, we utilize reward scores from all intermediate time steps to mitigate error accumulation over time.
In this ablation setting, however, the reward is computed solely based on the final output $\boldsymbol{x}_0$, which we refer to as ``Reward $\boldsymbol{x}_0$''.
By comparing ``Reward $\boldsymbol{x}_0$'' with the ``Original'' setting in Table~\ref{comparison1-abla}, we sho the advantage of leveraging rewards from multiple time steps rather than relying solely on the output of the final step.

\noindent\textbf{The effects of conducting RL with new normalization parameters.}
As described in Sec.~\ref{sec:diffusion}, we normalize the reward by computing the mean and variance using both the historical reward trajectory for each input image and the reward values from the current batch. For ablation, we consider a variant where the normalization parameters are computed using only the reward history of each input image. This setting is referred to as ``Norm. from track''. As shown in Table~\ref{comparison1-abla}, our proposed design offers clear advantages over this traditional normalization approach.

\noindent\textbf{The effects of conducting RL with different IQA rewards.}
In addition to using DeQA-Score as the reward model, we also evaluate the performance of our RL algorithm with alternative IQA models, including the MLLM-based Q-align and the traditional CLIP-IQA model. These ablation settings are referred to as ``with Q-align'' and ``with CLIP-IQA'', respectively. As shown in Table~\ref{comparison1-abla}, although their performance is lower than that of the ``Original'' setting with DeQA-Score, they still outperform the baselines. This suggests that stronger VLM-based IQA models might offer better performance, likely due to their superior generalization capabilities.

\noindent\textbf{Using iterative RL by updating the reward model.}
In this section, we conduct experiments by labeling the quality scores of the outputs from the diffusion models after training for several epochs. We then fine-tune the IQA model and apply RL once again. This setting is referred to as ``with Iter. RL''. As shown in Table~\ref{comparison1-abla}, the iterative application  of our method (with human in loop) leads to further improvements in performance.

%% file: sec/5_conclusion.tex
\section{Conclusion}

In this paper, we investigate an effective RL strategy for pre-trained diffusion-based restoration frameworks. We find that IQA metrics serve as effective reward functions, offering an alternative optimization direction for aligning with the distribution of high-quality images (e.g., ground truths)—distinct from the reconstruction-based objectives used in SFT.
To further enhance learning, we introduce difficulty-aware weighting, enabling RL to focus on hard samples—those that significantly differ from the ground truth. 
As the predictions increasingly align with the ground truth distribution, the optimization objective gradually shifts to incorporate SFT for fine-grained alignment.
Our method is plug-and-play and can be seamlessly integrated with any pre-trained diffusion model, boosting performance across various restoration tasks.